\documentclass[runningheads]{llncs}
\usepackage{graphicx}
\graphicspath{ {./figures/} }

\usepackage{url}
\usepackage{color}
\usepackage{ascmac}

\usepackage{amsmath}
\usepackage{amssymb}
\usepackage{amsfonts}
\usepackage{amsmath}
\usepackage{mathtools}

\usepackage{algorithmic}
\usepackage{algorithm}

\makeatletter
\newcommand{\defeq}{\mathrel{\aban@defeq}}
\newcommand{\aban@defeq}{%
  \vbox{\offinterlineskip\check@mathfonts
    \ialign{\hfil##\hfil\cr
      \fontsize{\ssf@size}{\z@}\normalfont def\cr
      \noalign{\kern1\p@}
      $\m@th=$\cr
      \noalign{\kern-.5\fontdimen22\textfont2}
    }%
  }%
}
\makeatother

\makeatletter
\newcommand{\encode}{\mathrel{\aban@encode}}
\newcommand{\aban@encode}{%
  \vbox{\offinterlineskip\check@mathfonts
    \ialign{\hfil##\hfil\cr
      \fontsize{\ssf@size}{\z@}\normalfont encode\cr
      \noalign{\kern1\p@}
      $\longrightarrow$\cr
      \noalign{\kern-.5\fontdimen22\textfont2}
    }%
  }%
}
\makeatother

\newcommand{\rem}[1]{\textit{\textcolor{magenta}{// {#1}}}}
\newcommand{\remc}[1]{\textit{\textcolor{magenta}{/* {#1} */}}}

\newcommand{\todo}[1]{\textit{\textcolor{red}{\if0 {#1} \fi}}}
\newcommand{\solved}[1]{\textit{\textcolor{magenta}{\if0 {#1} \fi}}}

\newcommand{\RQi}{\textbf{RQ1:} How much computation time is required to solve each problem?}
\newcommand{\RQii}{\textbf{RQ2:} What are the performance factors?}
\begin{document}
\title{Verifying Attention Robustness of Deep Neural Networks against Semantic Perturbations}
\titlerunning{Verifying Attention Robustness of DNNs against Semantic Perturbations}
\author{
  Satoshi Munakata\inst{1} \and
  Caterina Urban\inst{2} \and
  Haruki Yokoyama\inst{1} \and
  Koji Yamamoto\inst{1} \and
  Kazuki Munakata\inst{1}
}
\authorrunning{S. Munakata et al.}
\institute{
  Fujitsu, Kanagawa, Japan \and
  Inria \& ENS $\mid$ PSL \& CNRS, Paris, France
}
\maketitle              % typeset the header of the contribution
\begin{abstract}
\solved{Caterina: mentions \emph{combinations} of semantic perturbations, which I think is important) :}
It is known that deep neural networks (DNNs) classify an input image by paying particular attention to certain specific pixels; a graphical representation of the magnitude of attention to each pixel is called a \emph{saliency-map}. 
Saliency-maps are used to check the validity of the classification decision basis, e.g., it is not a valid basis for classification if a DNN pays more attention to the background rather than the subject of an image.
Semantic perturbations can significantly change the saliency-map. 
In this work, we propose the first verification method for \emph{attention robustness}, i.e., the local robustness of the changes in the saliency-map against combinations of semantic perturbations. Specifically, our method determines the range of the perturbation parameters (e.g., the brightness change) that maintains the difference between the actual saliency-map change and the expected saliency-map change below a given threshold value. Our method is based on activation region traversals, focusing on the outermost robust boundary for scalability on larger DNNs.
Experimental results demonstrate that our method can show the extent to which DNNs can classify with the same basis regardless of semantic perturbations and report on performance and performance factors of activation region traversals.

\keywords{
  feed-forward ReLU neural networks,
  robustness certification, 
  semantic perturbations, 
  saliency-map consistency,
  traversing
}
\end{abstract}

\todo{Caterina: I temporarly removed the ``if0'' from the todo/solved macros as I wanted to see them, sorry!}
\solved{Satoshi: Sure!}

%%%%%%%%%%%%%%%%%%%%%%%%%%%%%%%%%%%%%%%%%%%%%%%%%%%%%%%%%%%%%%%
% Introduction
%%%%%%%%%%%%%%%%%%%%%%%%%%%%%%%%%%%%%%%%%%%%%%%%%%%%%%%%%%%%%%%
\section{Introduction}

\noindent\textbf{Classification Robustness.}
% DNNは画像分類における支配的な手法である．
Deep neural networks (DNN) are dominant solutions in image classification \cite{Alex_NIPS_2012}.
% セーフティクリティカルシステムでは，DNNの品質保証が欠かせない．
However, quality assurance is essential when DNNs are used in safety-critical systems, for example, cyber-physical systems such as self-driving cars and medical diagnosis \cite{Rob_CSUR_2021}.
% 入力摂動に対する分類のロバスト性はキープロパティの1つである．
From an assurance point of view, the robustness of the classification against input perturbations is one of the key properties, and thus, it has been studied extensively \cite{Xiaowei_Computer_2020}.
% 通常のDNNは僅かなセマンティック摂動でも分類結果が変化してしまう．
\cite{Logan_ICML_2019,Xiang_Sensei_ICSE_2020} reported that despite input images can be perturbed in the real world by various mechanisms, such as \emph{brightness change} and \emph{translation}; even small \emph{semantic perturbations} can change the classification labels for DNNs.
% セマンティック摂動に対する分類のロバスト性の検証は重要．
Therefore, it is essential to verify classification robustness against semantic perturbations. 
Several methods have already been proposed to compute the range of perturbation parameters (e.g., the amount of brightness change and translation) that do not change classification labels \cite{Mislav_DeepG_NIPS_2019,Jeet_Semantify_CVPR_2020}. 

\noindent\textbf{Classification Validity.}
% DNNの注視領域は，分類結果の妥当性チェックに使われる．
It is known that DNNs classify an input image by paying particular attention to certain specific pixels in the image; a graphical representation of the magnitude of attention to each pixel, like a heatmap, is called \emph{saliency-map} \cite{Karen_ICLR_2014,Mukund_IG_ICML_2017}.
A saliency-map can be obtained from the gradients of DNN outputs with respect to an input image, and it is used to check the validity of the classification decision basis.
% 妥当性の低い分類結果は受け入れられない．
For instance, if a DNN classifies the subject type by paying attention to a background rather than the subject to be classified in an input image (as in the case of ``Husky vs. Wolf \cite{Marco_LIME_SIGKDD_2016}''), it is not a valid basis for classification.
We believe that such low validity classification should not be accepted in safety-critical situations, even if the classification labels are correct.
% セマンティック摂動は注視領域を変化させうる．
Semantic perturbations can significantly change the saliency-maps \cite{Gregoire_Digital_2018,Hao_CVPR_2019,Tao_ECT_AAAI_2021}.
% しかし，既存手法は注視領域の変化を検証できない．
However, existing robustness verification methods only target changes in the classification labels and not the saliency-maps. 

\noindent\textbf{Our Approach: Verifying Attention Robustness.}
% 本稿では，注視領域のロバスト性を検証する新手法を提案する．
In this work, we propose the first verification method for \emph{attention robustness}, i.e., the local robustness of the changes in the saliency-map against combinations of semantic perturbations.
% 本手法は注視領域が一貫性を保つ摂動許容範囲を算出する．
Specifically, our method determines the range of the perturbation parameters (e.g., the brightness change) that maintains the difference between (a) the actual saliency-map change and (b) the expected saliency-map change below a given threshold value.
% 例えば，明度変化は注視領域を変化させず，平行移動は注視領域を移動させる．
Regarding the latter (b), brightness change keeps the saliency-map unchanged, whereas translation moves one along with the image.
% 考え方は顕著性マップ一貫性と同じだけれど，点ではなく面で差を検証する必要がある．
Although the concept of such difference is the same as \emph{saliency-map consistency} used in semi-supervised learning \cite{Hao_CVPR_2019,Tao_ECT_AAAI_2021}, for the sake of verification, it is necessary to calculate the minimum and maximum values of the difference within each perturbation parameter sub-space.
% そこで，各活性区画内で差の最大値を算出する．
Therefore, we focus on the fact that DNN output is linear with respect to DNN input within an activation region \cite{Boris_NIPS_2019}.
That is, the actual saliency-map calculated from the gradient is constant within each region; thus, we can compute the range of the difference by sampling a single point within each region if the saliency-map is expected to keep, while by convex optimization if the saliency map is expected to move.
% 本手法は探索手法であり，ロバスト性の境界近傍のみを探索することもできる．
Our method is based on traversing activation regions on a DNN with layers for classification and semantic perturbations; it is also possible to traverse (i.e., verify) all activation regions in a small DNN or traverse only activation regions near the outermost robust boundary in a larger DNN.
% 実験から，提案手法の効果・性能・性能要因を示す．
Experimental results demonstrate that our method can show the extent to which DNNs can classify with the same basis regardless of semantic perturbations and report on performance and performance factors of activation region traversals.

\noindent\textbf{Contributions.}
Our main contributions are:
\begin{itemize}
% Attention Robustnessの検証問題を初めて定義し，検証手法を初めて提案した．
\item We formulate the problem of attention robustness verification; we then propose a method for verifying attention robustness for the first time. Using our method, it is also possible to traverse and verify all activation regions or only ones near the outermost decision boundary.
% 実装して性能を評価した．
\item We implement our method in a python tool and evaluate it on DNNs trained with popular datasets; we then show the specific performance and factors of verifying attention robustness. In the context of traversal verification methods, we use the largest DNNs for performance evaluation. 
\end{itemize}

%%%%%%%%%%%%%%%%%%%%%%%%%%%%%%%%%%%%%%%%%%%%%%%%%%%%%%%%%%%%%%%
% Overview
%%%%%%%%%%%%%%%%%%%%%%%%%%%%%%%%%%%%%%%%%%%%%%%%%%%%%%%%%%%%%%%
\section{Overview}

% In this section, we give an overview of our approach with motivating examples.

\begin{figure}
  % \caption{Motivating Examples.}
  \begin{tabular}{cc}
    \begin{minipage}[t]{0.95\hsize}
      \centering
      \includegraphics[keepaspectratio, scale=0.32]{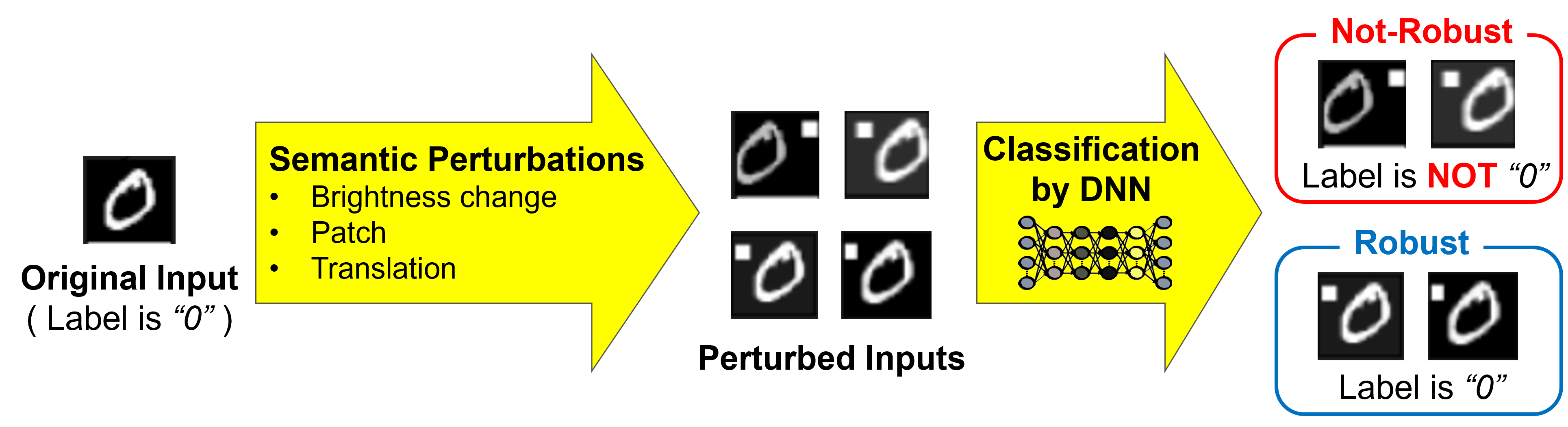}
      \caption{Misclassifications caused by combinations of semantic perturbations.}
      \label{fig:motivating_example_a}
    \end{minipage} \\

    \begin{minipage}[t]{0.95\hsize}
      \centering
      \includegraphics[keepaspectratio, scale=0.35]{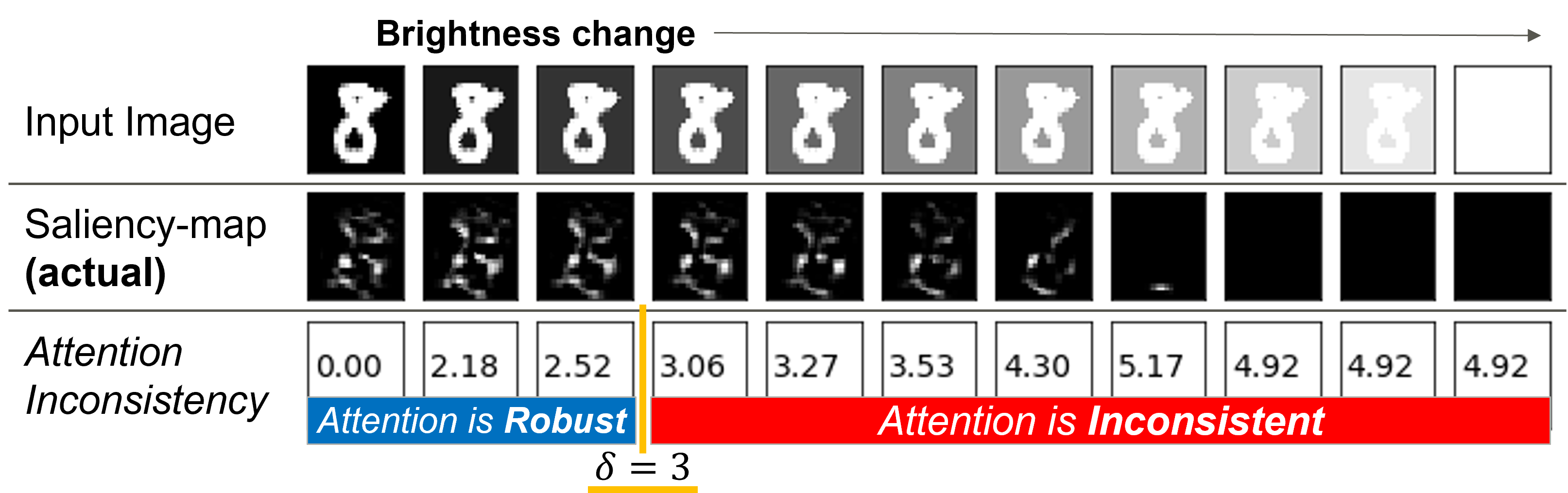}
      \caption{Perturbation-induced changes in images (first row), saliency-maps (second row) and the metric quantified the degree of collapse of each saliency-map (third row); where $\delta$ denotes the threshold to judge a saliency-map is valid or not.}
      \label{fig:motivating_example_b}
    \end{minipage} \\

    \begin{minipage}[t]{0.95\hsize}
      \centering
      \includegraphics[keepaspectratio, scale=0.35]{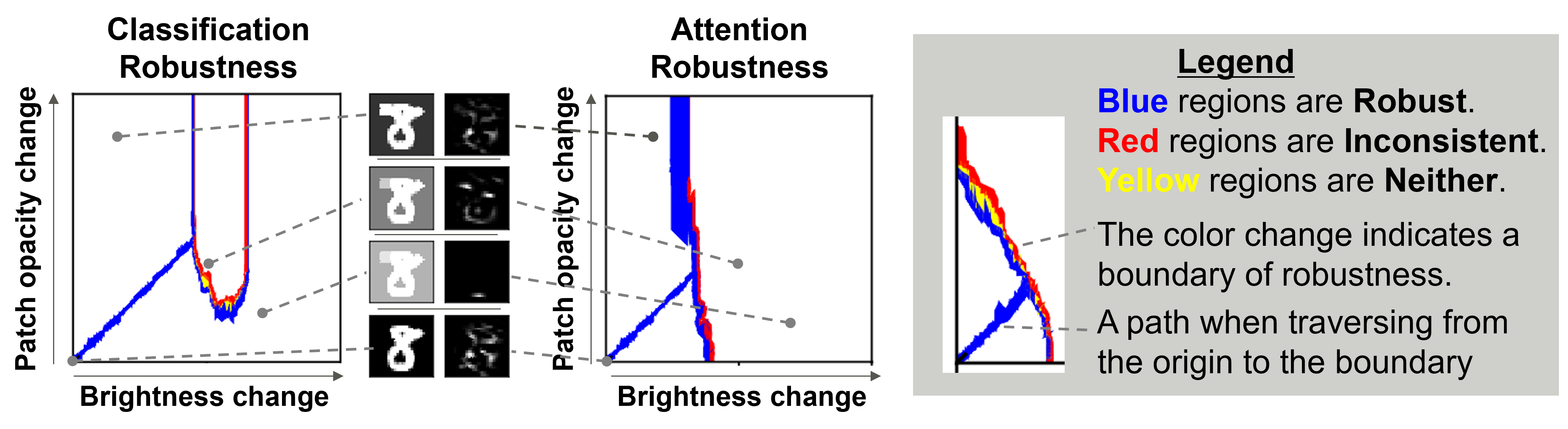}
      \caption{The outermost boundaries of classification robustness (left) and attention robustness (right); the origin at the bottom-left corresponds to the input image without perturbation, and each plotted point denotes the perturbed input image (middle). The shapes of the boundaries indicate the existence of regions that the DNN successfully classifies without sufficient evidence.}
      \label{fig:motivating_example_c}
    \end{minipage} \\\\

    \begin{minipage}[t]{0.95\hsize}
      \centering
      \includegraphics[keepaspectratio, scale=0.32]{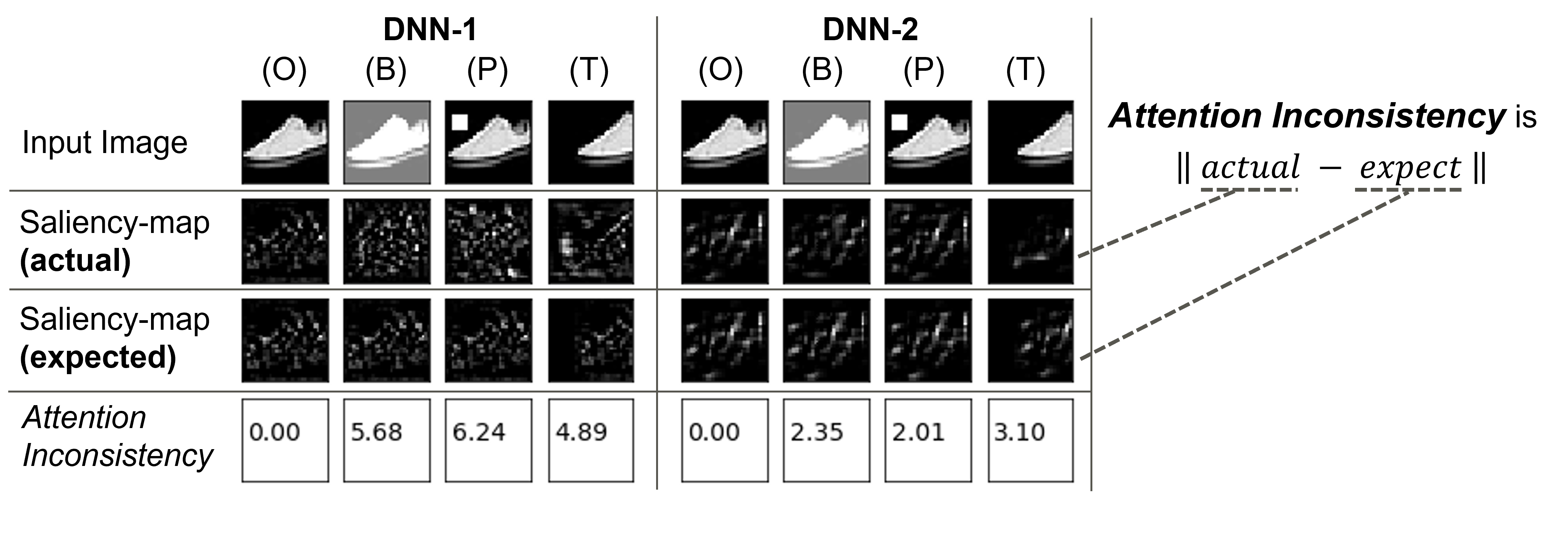}
      \caption{Differences in changes in saliency-maps for two DNNs. Each saliency-map of DNN-1 above is more collapsed than DNN-2's: where columns (O), (B), (P), and (T) denote original (i.e., without perturbations), brightness change, patch, and translation, respectively.}
      \label{fig:motivating_example_d}
    \end{minipage}

  \end{tabular}
\end{figure}

\noindent\textbf{Situation.}
% 画像分類用DNNのセマンティック摂動に対する弱さを評価したい．
Suppose a situation where we have to evaluate the weaknesses of a DNN for image classification against combinations of semantic perturbations caused by differences in shooting conditions, such as lighting and subject position.
% 例えば，明度変化や平行移動によりDNNは誤分類したりする．
For example, as shown in Figure~\ref{fig:motivating_example_a}, the original label of the handwritten text image is \textit{``0''}; however, the DNN often misclassifies it as the other labels, with changes in brightness, patch, and translations.
% なので，誤分類が起きやすい摂動範囲を把握しておきたい．
Therefore, we want to know in advance the ranges of semantic perturbation parameters that are likely to cause such misclassification as a weakness of the DNN for each typical image.
% しかし，分類ロバスト性だけでは不十分な場面がある．
However, classification robustness is not sufficient for capturing such weaknesses in the following cases.

% ケース1) 明度を変えても分類はロバストだけど，注目点は崩壊している
% * 明度変化後の注目点を見ると，オブジェクトに注目していないことがわかる．
% * 注目点の変化の大きさをaiとして定量化すると，閾値で妥当な範囲を絞り込める
% ケース2) 注目点が崩壊した画像は，パッチに弱くなっている
% * 摂動組合せに対するARの境界を可視化する
\noindent\textbf{Case 1.}
% 分類ロバスト性は偶然過大評価してしまうことがある．
Even if the brightness changes so much that the image is not visible to humans, the classification label of the perturbed image may happen to match the original label. 
Then vast ranges of the perturbation parameters are evaluated as robust for classification; however, such overestimated ranges are naturally invalid and unsafe.
% 図の1-2段目の説明．
For instance, Figure~\ref{fig:motivating_example_b} shows the changes in MNIST image ``8'' and the actual saliency-map when the brightness is gradually changed; although the classification seems robust because the labels of each image are the same, the collapsed saliency-maps indicate that the DNN does not pay proper attention to text ``8'' in each image.
% Attention Inconsistency, Attention Robustnessの説明．
Therefore, our approach uses the metric \emph{attention inconsistency}, which quantifies the degree of collapse of a saliency-map, to further evaluate the range of the perturbation parameter as satisfying the property \emph{attention robustness};
i.e., the DNN is paying proper attention as well as the original image.
Attention inconsistency is a kind of distance (cf. Figure~\ref{fig:motivating_example_d}) between an actual saliency-map (second row) and an expected one (third row);
e.g., the saliency-map of \textit{DNN-1} for translation perturbation (column (T)) is expected to follow image translation; however, if it is not, then attention inconsistency is high.
% 図の3段目の説明
In addition, Figure~\ref{fig:motivating_example_b} shows an example of determining that attention robustness is satisfied if each attention inconsistency value (third row) is less than or equal to threshold value $\delta$.

\noindent\textbf{Case 2.}
% 各摂動単体では耐えられた摂動範囲内でも，摂動を組合せるとしばしば誤分類する．
The classification label often changes by combining semantic perturbations, such as brightness change and patch, even for the perturbation parameter ranges that each perturbation alone could be robust.
%% 想定されるセマンティック摂動は多数あるので，全組合せは検証困難．
It is important to understand what combinations are weak for the DNN; however, it is difficult to verify all combinations as there are many semantic perturbations assumed in an operational environment.
% 我々の観測では，顕著性マップを崩壊させる摂動は，組合せに弱い．
In our observations, a perturbation that significantly collapses the saliency-map is more likely to cause misclassification when combined with another perturbation because another perturbation can change the intensity of pixels to which the DNN should not pay attention.
% Outermost boundaryの説明．
Therefore, to understand the weakness of combining perturbations, our approach visualizes the \emph{outermost boundary} at which the sufficiency of robustness switches on the perturbation parameter space.
% 図の説明．
For instance, Figure~\ref{fig:motivating_example_c} shows connected regions that contain the outermost boundary for classification robustness (left side) or attention robustness (right side).
The classification boundary indicates that the DNN can misclassify the image with a thin patch and middle brightness. 
In contrast, the attention boundary further indicates that the brightness change can collapse the saliency-map more than patching, so we can see that any combinations with the brightness change pose a greater risk.
% 図の説明．
Even when the same perturbations are given, the values of attention inconsistency for different DNNs are usually different (cf. Figure~\ref{fig:motivating_example_d}); thus, it is better to evaluate what semantic perturbation poses a greater risk for each DNN.

%%%%%%%%%%%%%%%%%%%%%%%%%%%%%%%%%%%%%%%%%%%%%%%%%%%%%%%%%%%%%%%
% Problem Formulation
%%%%%%%%%%%%%%%%%%%%%%%%%%%%%%%%%%%%%%%%%%%%%%%%%%%%%%%%%%%%%%%
\section{Problem Formulation} \label{sec:problem}

Our method targets feed-forward ReLU-activated neural networks (ReLU-FNNs) for image classification. A ReLU-FNN \emph{image classifier} is a function  $f\colon X \to Y$ mapping an $N^f$-dimensional (pixels $\times$ color-depth) image $x \in X \subseteq \mathbb{R}^{N^f}$ to a classification label $argmax_{j \in Y} f_j(x)$ in the  $K^f$-class label space $Y  = \{1,\dots,K^f\}$, where $f_j: X \to \mathbb{R}$ is the confidence function for the j-th class.

The ReLU activation function occurs in between the linear maps performed by the ReLU-FNN layers and applies the function $max(0, x_{l, n})$ to each neuron $x_{l, n}$ in a layer $l \in L^f$ (where $L^f$ is the number of layers of ReLU-FNN $f$). When $x_{l, n} > 0$, we say that $x_{l, n}$ is \emph{active}; otherwise, we say that $x_{l, n}$ is \emph{inactive}. We write $ap^f(x)$ for the \emph{activation pattern} of an image $x$ given as input to a ReLU-FNN $f$, i.e., the sequence of neuron activation statuses in $f$ when $x$ is taken as input. We write $AP^f$ for the entire set of activation patterns of a ReLU-FNN $f$.

Given an activation pattern $p \in AP^f$, we write $ar^f(p)$ for the corresponding \emph{activation region}, i.e., the subset of the input space containing all images that share the same activation pattern: $x \in ar^f(p) \Leftrightarrow ap^f(x) = p$. Note that, neuron activation statuses in an activation pattern $p$ yield half-space constraints in the input space \cite{Boris_NIPS_2019,Matt_GeoCert_NIPS_2019}. Thus, an activation region $ar^f(p)$ can equivalently be represented as a convex polytope described by the conjunction of the half-space constraints resulting from the activation pattern $p$.

\subsubsection{Classification Robustness.} \label{subsec:cr}

A \emph{semantic perturbation} is a function $g: \Theta \times X \to X$ applying a perturbation with $N^g$ parameters $\theta \in \Theta \subseteq \mathbb{R}^{N^g}$ to an image $x \in X$ to yield a perturbed image \todo{Satoshi: Does $g_g$ mean $g_{N^g}$?} $g(\theta,x) \defeq g_{N^g}(\theta_{N^g},\cdot) \circ \dots \circ g_1(\theta_1,x) = g_{N^g}(\theta_{N^g},\dots g_1(\theta_1,x), \dots) \in X$, where $g_i: \mathbb{R} \times X \to X$ performs the $i$-th atomic semantic perturbation with parameter $\theta_i$ (with $g_i(0, x) = x$ for any image $x \in X$). For instance, a brightness decrease perturbation $g_b$ is a(n atomic) semantic perturbation function with a single brightness adjustment parameter $\beta \geq 0$: $g_b (\beta, x) \defeq ReLU(x - \vec{1}\beta)$.

\begin{definition}[Classification Robustness] 
A perturbation region $\eta \subset \Theta$ satisfies \emph{classification robustness} --- written $CR(x;\eta)$ --- if and only if the classification label $f(g(\theta,x))$ is the same as $f(x)$ when the perturbation parameter $\theta$ is within $\eta$: $CR(x;\eta) \defeq \forall \theta \in \eta.\; f(x) = f(g(\theta,x))$.\end{definition}

Vice versa, we define \emph{misclassification robustness} when $f(g(\theta,x))$ is always different from $f(x)$ when $\theta$ is within $\eta$: $MR(x;\eta) \defeq \forall \theta \in \eta.\; f(x) \neq f(g(\theta,x))$.
\solved{Note that I changed 'classification violateness' into 'misclassification robustness' (and thus 'CV' into 'MR', mostly because 'violateness' does not seem to be an existing English word ;)}
\solved{Satoshi: I see. I will use 'misclassification robustness' and 'MR' in this paper.}

The \emph{classification robustness verification problem} $Prob^{CR} \defeq (f, g, x0, \Theta)$ consists in enumerating, for a given input image $x0$, the perturbation parameter regions $\eta^{CR}, \eta^{MR} \subset \Theta$ respectively satisfying $CR(x0;\eta^{CR})$ and $MR(x0;\eta^{MR})$.

\subsubsection{Attention Robustness.} \label{subsec:ar}

We generalize the definition of saliency-map from \cite{Karen_ICLR_2014} to that of an \emph{attention-map}, which is a function $map_j: X \to X$ from an image $x \in X$ to the heatmap image $m_j \in X$ plotting the magnitude of the contribution to the j-th class confidence $f_j(x)$ for each pixel of $x$. Specifically, $map_j(x) \defeq filter\Bigl(\frac{\partial f_j(x)}{\partial x_1}, \dots, \frac{\partial f_j(x)}{\partial x_{N^f}}\Bigr)$, where $filter(\cdot)$ is an arbitrary image processing function (such as normalization and smoothing) and, following \cite{Karen_ICLR_2014,Dimitris_ICLR_2019}, the magnitude of the contribution of each pixel $x_1, \dots, x_{N^f}$ is given by the gradient with respect to the $j$-th class confidence. When $filter(x) \defeq |x|$, our definition of $map_j$ matches that of saliency-map in \cite{Karen_ICLR_2014}. Note that, within an activation region $ar^f(p)$, $f_j$ is linear  \cite{Boris_NIPS_2019} and thus the gradient $\frac{\partial f_j(x)}{\partial x_i}$ is a constant value.

We expect attention-maps to change consistently with respect to a semantic image perturbation. For instance, for a brightness change perturbation, we expect the attention-map to remain the same. Instead, for a translation perturbation, we expect the attention-map to be subject to the same translation. In the following, we write $\tilde{g}(\cdot)$ for the attention-map perturbation corresponding to a given semantic perturbation $g(\cdot)$. We define \emph{attention inconsistency} as the difference between the actual and expected attention-map after a semantic perturbation:
\[
  ai(x;\theta) \defeq \sum_{j \in Y} dist\Bigl( map_j\bigl(g(\theta,x)\bigr), \tilde{g}\bigl(\theta,map_j(x)\bigr) \Bigr) 
\]
where $dist\colon X \times X \to \mathbb{R}$ is an arbitrary distance function such as Lp-norm ($||x - x'||_p$). Note that, when $dist(\cdot)$ is L2-norm, our definition of attention inconsistency coincides with the definition of saliency-map consistency given by \cite{Hao_CVPR_2019}.

\begin{definition}[Attention Robustness] 
A perturbation region $\eta \subset \Theta$ satisfies \emph{attention robustness} --- written $AR(x;\eta,\delta)$ --- if and only if the attention inconsistency is always less than or equal to $\delta$ when the perturbation parameter $\theta$ is within $\eta$: $AR(x;\eta,\delta) \defeq \forall \theta \in \eta.\; ai(x;\theta) \leq \delta$.\end{definition}

When the attention inconsistency is always greater than $\delta$, we have \emph{inconsistency robustness}: $IR(x;\eta,\delta) \defeq \forall \theta \in \eta.\; ai(x;\theta) > \delta$.

The \emph{attention robustness verification problem} $Prob^{AR} \defeq (f, g, x0, \Theta, \delta)$ consists in enumerating, for a given input image $x0$, the perturbation parameter regions $\eta^{AR}, \eta^{IR} \subset \Theta$ respectively satisfying $AR(x0;\eta^{AR},\delta)$ and $IR(x0;\eta^{IR},\delta)$.

\subsubsection{Outermost Boundary Verification.} \label{subsec:ob}

\begin{figure}[t]
  \centering
  \includegraphics[width=30em]{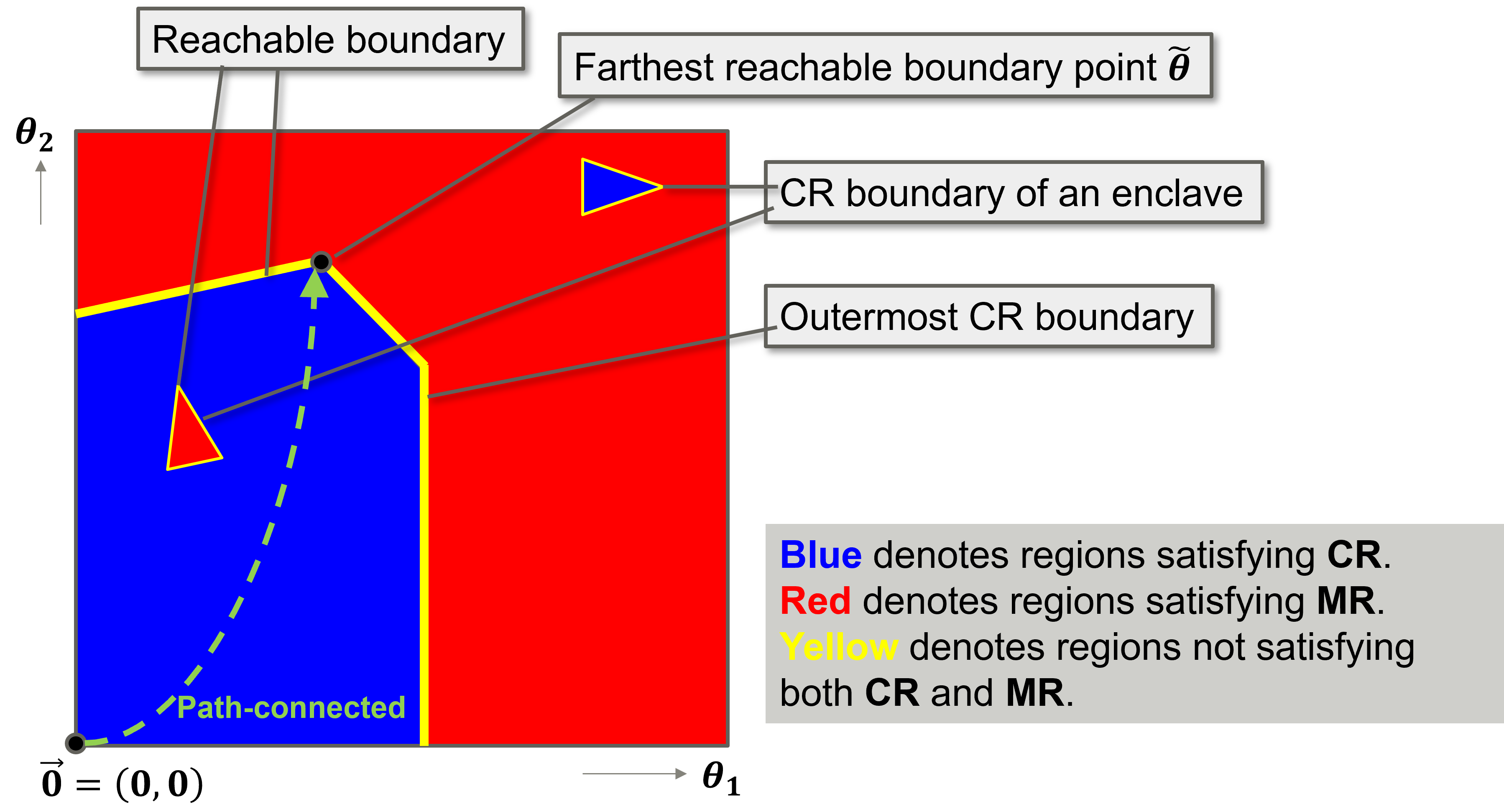}
  \caption{Illustration of outermost CR boundary on a 2-dimensional perturbation parameter space. The origin $\vec{0}$ is the original image without perturbation ($\theta_1 = \theta_2 = 0$).}
  \label{fig:ob}
\end{figure}

In practice, to represent the trend of the weakness of a ReLU-FNN image classifier to a semantic perturbation, we argue that it is not necessary to enumerate all perturbation parameter regions within a perturbation parameter space $\Theta$. Instead, we search the \emph{outermost $CR$/$AR$ boundary}, that is, the perturbation parameter regions $\eta$ that lay on the $CR$/$AR$ boundary farthest away from the original image. 

An illustration of the outermost CR boundary is given in Figure~\ref{fig:ob}. More formally, we define the outermost $CR$ boundary as follows:

\begin{definition}[Outermost $CR$ Boundary] 
The outermost $CR$ boundary of a classification robustness verification problem, $ob^{CR}(Prob^{CR})$, is a set of perturbation parameter regions $HS \subset \mathcal{P}(\Theta)$ such that:
\begin{enumerate}
%\item $HS$ forms a connected-space, i.e., a topological space that cannot be represented as the union of two or more disjoint non-empty open subsets;
\item for all perturbation regions $\eta \in HS$, there exists a path connected-space from the original image $x0$ (i.e., $\vec{0} \in \Theta$) to $\eta$ that consists of regions satisfying $CR$ (written $Reachable(\eta; x0)$);
\item all perturbation regions $\eta \in HS$ lay on the classification boundary, i.e., $\exists \theta,\theta' \in \eta.\; f(g(\theta,x0)) = f(x0) \land f(g(\theta',x0)) \neq f(x0)$;
\item there exists a region $\eta \in HS$ that contains the farthest reachable perturbation parameter point $\tilde{\theta}$ from the original image, i.e., $\tilde{\theta} = max_{\theta \in \Theta} ||\theta||_2$ such that $Reachable(\{\theta\}; x0)$.
\end{enumerate}
\end{definition}
The definition of the outermost $AR$ boundary is analogous. Note that not all perturbation regions inside the outermost $CR$/$AR$ boundary satisfy the $CR$/$AR$ property (cf. the enclaves in Figure~\ref{fig:ob}).

The \emph{outermost $CR$ boundary verification problem} and \emph{outermost $AR$ boundary verification problem} $Prob^{CR}_{ob} = (f, g, x0, \Theta)$ and $Prob^{AR}_{ob} = (f, g, x0, \Theta, \delta)$ consist in enumerating, for a given input image $x0$, the perturbation parameter regions $\eta^{CR}_{ob}$ and $\eta^{AR}_{ob}$) that belong to the outermost $CR$ and $AR$ boundary $ob^{CR}(Prob^{CR})$ and $ob^{AR}(Prob^{AR})$.

%%%%%%%%%%%%%%%%%%%%%%%%%%%%%%%%%%%%%%%%%%%%%%%%%%%%%%%%%%%%%%%
% Method
%%%%%%%%%%%%%%%%%%%%%%%%%%%%%%%%%%%%%%%%%%%%%%%%%%%%%%%%%%%%%%%
\section{Geometric Boundary Search (\textit{GBS})}

In the following, we describe our Geometric Boundary Search (\textit{GBS}) method for solving $Prob^{CR}_{ob}$, and $Prob^{AR}_{ob}$ shown in Algorithm~\ref{alg:gbs} and~\ref{alg:gbs_leftover}.
In Appendix~\ref{sec:bfs}, we also describe our Breadth-First Search (\textit{BFS}) method for solving $Prob^{CR}$, and $Prob^{AR}$.

%%%%%%%%%%%%%%%%%%%%%%%%%%%%%%%%%%%%%%%
\subsection{Encoding Semantic Perturbations} \label{sec:encode_sp}

After some variables initialization (cf. Line~\ref{line:init} in Algorithm~\ref{alg:gbs}), the semantic perturbation $g$ is encoded into a ReLU-FNN $g^{x0}: \Theta \to X$ (cf. Line~\ref{line:g}). 

In this paper, we focus on combinations of atomic perturbations such as brightness change (B), patch placement (P), and translation (T). 
Nonetheless, our method is applicable to any semantic perturbation as long as it can be represented or approximated with sufficient accuracy. 
% For instance, our method is not applicable to a semantic perturbation $g$ for which a corresponding saliency-map change $\tilde{g}$ cannot be defined and computed, e.g., DNNs that generate an image from an input text (such as \cite{Ramesh_DALL_ICML_2021}).

For the encoding, we follow \cite{Jeet_Semantify_CVPR_2020} and represent (combinations of) semantic perturbations as a piecewise linear function by using affine transformations and ReLUs. For instance, a brightness decrease perturbation $g_b (\beta, x0) \defeq ReLU(x0 - \vec{1}\beta)$ (cf. Section~\ref{subsec:cr}) can be encoded as a ReLU-FNN as follows:
\begin{equation*}
 g_b(\beta, x0) \encode     \begin{bmatrix}
      1 & 0 & \dots & 0 \\
      0 & 1 & \dots & 0 \\
         &   & \dots &   \\
      0 & 0 & \dots & 1 \\
    \end{bmatrix} \: ReLU \left( 
 \begin{bmatrix}
      -1 & 1 & 0 & \dots & 0 \\
      -1 & 0 & 1 & \dots & 0 \\
         &   &   & \dots &   \\
      -1 & 0 & 0 & \dots & 1 \\
    \end{bmatrix}
    \begin{bmatrix}
      \beta \\
      x0_1 \\
      \dots \\
      x0_{N^f} \\
    \end{bmatrix}
  \right) + \vec{0}
\end{equation*}
which we can combine with the given ReLU-FNN $f$ to obtain the compound ReLU-FNN $f \circ g_b^{x0}$ to verify.
The full encoding for all considered (brightness, patch, translation) perturbations is shown in Appendix~\ref{sec:encode_sp_full}. \todo{shown in the Appendix. @Satoshi: Could you add the full encoding to the Appendix?} \todo{Satoshi: I see. I will add it.}

%%%%%%%%%%%%%%%%%%%%%%%%%%%%%%%%%%%%%%%
\subsection{Traversing Activation Regions.}

\textit{GBS} then performs a traversal of activation regions of the compound ReLU-FNN $f \circ g^{x0}$ near the outermost $CR$/$AR$ boundary for $Prob^{CR}_{ob}$/$Prob^{AR}_{ob}$.
Specifically, it initializes a queue $Q$ with the activation pattern $ap^{f \circ g^{x0}}(\vec{0})$ of the original input image $x0$ with no semantic perturbation, i.e., $\theta = \vec{0}$ (cf. Line~\ref{line:gbs_q} in Algorithm~\ref{alg:gbs}, we explain the other queue initialization parameters shortly). Given a queue element $q \in Q$, the functions $p(q)$, $isFollowing(q)$, and $lineDistance(q)$ respectively return the 1st, 2nd, and 3rd element of $q$.

\begin{figure}[t]
  \centering
  \includegraphics[width=\linewidth]{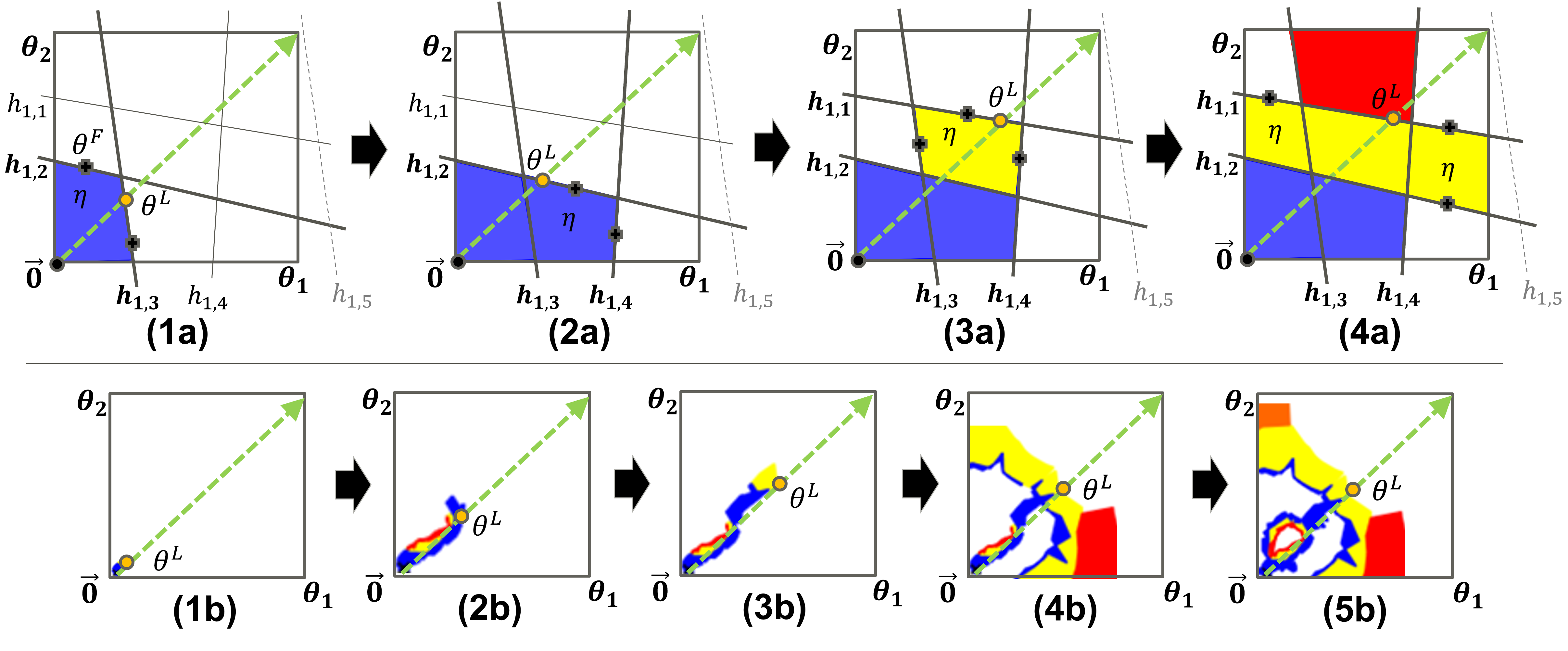}
  \caption{A running example of \textit{GBS}. The upper row shows the basic traversing flow, while the lower row shows the flow of avoiding enclaves. $h_{l,n}$ denotes a half-space corresponding to neuron activity $p_{l,n}$.}
  \label{fig:running_example_a}
\end{figure}

Then, for each activation pattern $p$ in $Q$ (cf. Line~\ref{line:gbs_p}), \textit{GBS} reconstructs the corresponding perturbation parameter region $\eta$ (subroutine \emph{constructActivationRegion}, Line~\ref{line:car}) as the convex polytope resulting from $p$ (cf. Section~\ref{sec:problem} and $\eta$ in Figure~\ref{fig:running_example_a}-(1a)). 

Next, for each neuron $x_{l,n}$ in $f \circ g^{x0}$ (cf. Line~\ref{line:gbs_for}), it checks whether its activation status cannot flip within the perturbation parameter space $\Theta$, i.e., the resulting half-space has no feasible points within $\Theta$ (subroutine \emph{isStable}, Line~\ref{line:gbs_stable}, cf. half-space $h_{1,5}$ in Figure~\ref{fig:running_example_a}-(1a)). 
Otherwise, a new activation pattern $p'$ is constructed by flipping the activation status of $x_{l,n}$ (subroutine \emph{flipped}, Line~\ref{line:gbs_flip}) and added to a local queue $Q'$ (cf. Line~\ref{line:gbs_local}, and ~\ref{line:gbs_add2},~\ref{line:gbs_add3}) if $p'$ has not been observed already (cf. Line~\ref{line:gbs_new}) and it is feasible (subroutine \emph{calcInteriorPointOnFace}, Lines~\ref{line:gbs_face}-\ref{line:gbs_feasible}. cf. point $\theta^F$ and half-space $h_{1,2}$ in Figure~\ref{fig:running_example_a}-(1a)).

The perturbation parameter region $\eta$ is then simplified to $\tilde{\eta}$ (subroutine \emph{simplified}, Line~\ref{line:gbs_simplify} in Algorithm~\ref{alg:gbs_leftover}; e.g., reducing the half-spaces used to represent $\eta$ to just $h_{1,2}$ and $h_{1,3}$ in Figure~\ref{fig:running_example_a}-(1a)).
$\tilde{\eta}$ is used to efficiently calculate the range of attention inconsistency within $\eta$ (subroutine \emph{calcRange}, Line~\ref{line:gbs_calcRange} in Algorithm~\ref{alg:gbs_leftover}, cf. Section~\ref{sec:verify}), and then attention/inconsistency robustness can be verified based on the range (Line~\ref{line:gbs_ar} and~\ref{line:gbs_ir} in Algorithm~\ref{alg:gbs_leftover}).
Furthermore, classification/misclassification robustness can be verified in the same way if subroutine $calcRange$ returns the range of confidence $f_{f(x0)}(g^{x0}(\theta)) - f_j(g^{x0}(\theta))$ within $\tilde{\eta}$ (cf. Section~\ref{sec:verify}) and $\delta = 0 \land w^\delta = 0$.
At last, the local queue $Q'$ is pushed onto $Q$ (cf. Line~\ref{line:gbs_push} in Algorithm~\ref{alg:gbs}).

To avoid getting stuck around enclaves inside the outermost $CR$/$AR$ boundary (cf. Figure~\ref{fig:ob}) during the traversal of activation regions, \textit{GBS} switches status when needed between ``searching for a decision boundary'' and ``following a found decision boundary''. 
The initial status is set to ``searching for a decision boundary'', i.e., $\neg isFollowing$ when initializing the queue $Q$ (cf. Line~\ref{line:gbs_q}).
The switch to $isFollowing$ happens when region $\eta$ is on the boundary (i.e., $lo \leq \delta \leq up$) or near the boundary (i.e., $\delta - w^\delta \leq lo \leq \delta + w^\delta$, cf. Line~\ref{line:gbs_switch} in Algorithm~\ref{alg:gbs_leftover} and Figure~\ref{fig:running_example_a}-(3a,1b,3b)); where $w^\delta$ is a hyperparameter to determine whether close to the boundary or not. 
$w^\delta$ should be greater than $0$ to verify attention/inconsistency robustness because attention inconsistency changes discretely for ReLU-FNNs (ch. Section~\ref{sec:calc_ac}).
\textit{GBS} can revert back to searching for a decision boundary if when following a found boundary it finds a reachable perturbation parameter region that is farther from $\vec{0}$ (cf. Lines~\ref{line:gbs_distance}-\ref{line:gbs_back} in Algorithm~\ref{alg:gbs} and Figure~\ref{fig:running_example_a}-(2b)).

%%%%%%%%%%%%%%%%%%%%%%%%%%%%%%%%%%%%%%%
% Algorithm GBS．
%%%%%%%%%%%%%%%%%%%%%%%%%%%%%%%%%%%%%%%
\begin{algorithm}[t]
  \caption{$gbs(f,g,x0,\Theta;\delta,w) \to (H^{CR},H^{MR},H^{CB}, H^{AR},H^{IR},H^{AB})$}
  \label{alg:gbs}

  \begin{algorithmic}[1]
    % Input/Output
    \renewcommand{\algorithmicrequire}{\textbf{Input:}}
    \renewcommand{\algorithmicensure}{\textbf{Output:}}
    \REQUIRE $f, g, x0, \Theta, \delta$
    \ENSURE  $H^{CR},H^{MR},H^{CB}, H^{AR},H^{IR},H^{AB} \subset \mathcal{P}(\Theta)$

    % Statements
    \STATE $H^{CR},H^{MR},H^{CB}, H^{AR},H^{IR},H^{AB} \leftarrow \{\},\{\},\{\},\{\},\{\},\{\}$\label{line:init}
    \STATE $g^{x0} \leftarrow g(\cdot,x0)$ \rem{currying $g$ with $x0$; i.e, $g^{x0}(\theta) = g(\theta,x0)$.} \label{line:g}
    \STATE $Q \subset AP^{f \circ g^{x0}} \times \mathbb{B} \times \mathbb{R} \leftarrow \{( ap^{f \circ g^{x0}}(\vec{0}), \bot, 0 )\}$ \rem{queue for boundary search.} \label{line:gbs_q}
    % \STATE %\;\;\;\;\;\;\;\; 
    \STATE $OBS \subset AP^{f \circ g^{x0}} \leftarrow \{\}$ \rem{observed activation patterns.}

    \WHILE {$\#|Q| > 0$ \rem{loop for geometrical-boundary search.}}
      \STATE $q \leftarrow popMaxLineDistance(Q)$; $p \leftarrow p(q)$\label{line:gbs_p}; $OBS \leftarrow OBS \cup \{p\}$
      \STATE $\eta \leftarrow constructActivationRegion(f \circ g^{x0}, p)$\label{line:car}
      \STATE $FS \subset \mathbb{Z} \times \mathbb{Z} \leftarrow \{\} $ \rem{$(l,n)$ means the $n$-th neuron in $l$-th layer is a face of $\eta$}
      \STATE $Q' \subset AP^{f \circ g^{x0}} \times \mathbb{B} \times \mathbb{R} \leftarrow \{\}$ \rem{local queue for an iteration.} \label{line:gbs_local}

      \STATE \rem{Push each activation region connected to $\eta$.}
      \FOR {$l = 1$ \TO \#layers of $f \circ g^{x0}$; $n = 1$ \TO \#neurons of the $l$-th layer} \label{line:gbs_for}
        \STATE continue \textbf{if} $isStable(p,l,n,\Theta)$ \rem{skip if activation of $x_{l,n}$ cannot flip in $\Theta$.} \label{line:gbs_stable}

        \STATE $p' \leftarrow flipped(p,l,n)$ \rem{flip activation status for neuron $x_{l,n}$.} \label{line:gbs_flip}
        \STATE continue \textbf{if} {$p' \in OBS$} \textbf{else} $OBS \leftarrow OBS \cup \{p'\}$ \rem{skip if $p'$ was observed.} \label{line:gbs_new}

        \STATE $\theta^F \leftarrow calcInteriorPointOnFace(\eta,l,n)$ \label{line:gbs_face}
        \STATE continue \textbf{if} {$\theta^F = null$} \rem{skip if $p'$ is infeasible.} \label{line:gbs_feasible}
        \STATE $FS \leftarrow FS \cup \{(l,n)\}$ \rem{(l,n) is a face of $\eta$.}

        \STATE $\theta^L \leftarrow calcInteriorPointOnLine(\eta,l,n)$\label{line:gbs_line}
        \IF {$isFollowing(q) \land \theta^L \neq null \land ||\theta^L||_2 > lineDistance(q)$} \label{line:gbs_distance}
          \STATE $q \leftarrow (p, \bot, lineDistance(q))$ \rem{Re-found the line in boundary-following.} \label{line:gbs_back}
        \ENDIF
        
        \IF {$\lnot isFollowing(q) \land \theta^L \neq null$}
          \STATE $Q' \leftarrow Q' \cup \{( p', \bot, ||\theta^L||_2 )\}$ \rem{continue line-search.} \label{line:gbs_add2}
        \ELSE
          \STATE $Q' \leftarrow Q' \cup \{( p', isFollowing(q), lineDistance(q) )\}$ \rem{continue current.} \label{line:gbs_add3}
        \ENDIF
      \ENDFOR

      \STATE \textit{\textbf{(...Verify $\eta$...)}} \rem{See Algorithm~\ref{alg:gbs_leftover} for $AR$/$IR$ (analogous for $CR$/$MR$)}

      \STATE $Q \leftarrow Q \cup Q'$ \rem{Push} \label{line:gbs_push}
    \ENDWHILE
  \end{algorithmic}
\end{algorithm}

\begin{algorithm}
  \caption{(Expanding from Algorithm~\ref{alg:gbs} for $AR(x0;\eta,\delta)$/$IR(x0;\eta,\delta)$)}
  \label{alg:gbs_leftover}
  \begin{algorithmic}[1]
    % Statements
    \STATE \textit{\textbf{(...Verify $\eta$...)}}
    \STATE $\tilde{\eta} \leftarrow simplified(\eta,FS)$ \rem{limit the constraints on $\eta$ to $FS$.} \label{line:gbs_simplify}
    \STATE $(lo, up) \leftarrow calcRange(x0;\tilde{\eta})$ \rem{the range ([lower and upper) of ai within $\tilde{\eta}$.} \label{line:gbs_calcRange}
    \STATE $nearBoundary \leftarrow (lo \leq \delta \leq up) \lor (\delta - w^\delta \leq lo \leq \delta + w^\delta) \lor (\delta - w^\delta \leq up \leq \delta + w^\delta)$ \label{line:gbs_nearBoundary}

    \IF {$lo \leq up \leq \delta$ \remc{satisfying AR}} \label{line:gbs_ar}
      \STATE $H^{AR} \leftarrow H^{AR} \cup \{\tilde{\eta}\}$
      \STATE $Q' \leftarrow \{\}$ \textbf{if} $isFollowing(q) \land \lnot nearBoundary$ \rem{no traversing connected regions.}
    \ELSIF {$\delta < lo \leq up$  \remc{satisfying IR}} \label{line:gbs_ir}
      \STATE $H^{IR} \leftarrow H^{IR} \cup \{\tilde{\eta}\}$
      \STATE $Q' \leftarrow \{\}$ \textbf{if} $\lnot nearBoundary$ \rem{no traversing connected regions.}
    \ELSE
      \STATE $H^{AB} \leftarrow H^{AB} \cup \{\tilde{\eta}\}$
    \ENDIF

    \IF {$\lnot isFollowing(q) \land nearBoundary$}
      \STATE \textit{\textbf{(...Update $Q'$ such that $\forall q' \in Q.\; isFollowing(q')$...)}} \label{line:gbs_switch} \rem{switch to boundary-following.}
    \ENDIF
  \end{algorithmic}
\end{algorithm}

%%%%%%%%%%%%%%%%%%%%%%%%%%%%%%%%%%%%%%%
\subsection{Calculating Attention Inconsistency} \label{sec:calc_ac}

% 勾配は定数．
\noindent\textbf{Gradients within an Activation Region.}
Let $p \in AP^{f \circ g^{x0}}$ be an activation pattern of the compound ReLU-FNN $f \circ g^{x0}$. The gradient $\frac{\partial f_j(g^{x0}(\theta))}{\partial \theta_s}$ is constant within $ar^{f \circ g^{x0}}(p)$ (cf. Section~\ref{subsec:ar}). We write $g_i^{x0}(\theta)$ for the $i$-th pixel $x_i$ of a perturbed image in $\{ g^{x0}(\theta) \mid \theta \in ar^{f \circ g^{x0}}(p) \} \subset X$. The gradient $\frac{\partial g_i^{x0}}{\partial \theta} = \frac{\partial x_i}{\partial \theta}$ is also a constant value. By the chain rule, we have $\frac{\partial f_j(x)}{\partial x_i} = \frac{\partial f_j(g^{x0}(\theta)) / \partial \theta_s}{\partial x_i / \theta_s}$. Thus $\frac{\partial f_j(x)}{\partial x_i}$ is also constant. This fact is formalized by the following lemma:
\begin{lemma} \label{eq:lemma1}
  $\frac{\partial f_j(x)}{\partial x_i} = C \;\; (x \in \{ g^{x0}(\theta) \mid \theta \in ar^{f \circ g^{x0}}(p) \})$
\end{lemma}
(cf. a small example in Appendix~\ref{eg:lemma1}).
%
% f(g(・))の活性区画内での勾配は，f(x)の下流の活性パターンが一致する活性区画の勾配と一致する．
Therefore, the gradient $\frac{\partial f_j(x)}{\partial x_i}$ can be computed as the weights of the j-th class output for ReLU-FNN $f$ about activation pattern $ap^f(\dot{x})$; where, $\dot{x} = g^{x0}(\dot{\theta})$ and $\dot{\theta}$ is an arbitrary sample within $ar^{f \circ g^{x0}}(p)$ (cf. Appendix~\ref{sec:linearity}).

% 勾配の摂動は，gを勾配についてカリー化することでReLU-FNNに変換する．
% \noindent\textbf{ReLU-FNN of Perturbed Gradients.}
For the perturbed gradient $\tilde{g}(\theta,\frac{\partial f_j(x)}{\partial x_i})$, let $\tilde{g}(\theta)$ be the ReLU-FNN $g^{\frac{\partial f_j(x)}{\partial x_i}}(\theta')$.
Thus, the same consideration as above applies.

% filterとdistが凸関数なら凸最適化が使える．
\noindent\textbf{Attention Inconsistency (ai).}
We assume both $filter(\cdot)$ and $dist(\cdot)$ are convex downward functions for calculating the maximum / minimum value by convex optimization.
Specifically, $filter(\cdot)$ is one of the identity function (I), the absolute function (A), and the $3 \times 3$ mean filter (M).
$dist(\cdot)$ is one of the $L_1$-norm ($L_1$) and the $L_2$-norm ($L_2$): where, $w$ is the width of image $x \in X$.

%%%%%%%%%%%%%%%%%%%%%%%%%%%%%%%%%%%%%%%
\subsection{Verifying CR and AR within an activation region} \label{sec:verify}

Our method leverages the fact that the gradient of a ReLU-FNN output with respect to the input is constant within an activation region (cf. Section~\ref{subsec:ar}); thus, $CR/MR$ can be resolved by linear programming, and $AR/IR$ can be resolved by just only one sampling if the saliency-map is expected to keep or convex optimization if the saliency-map is expected to move.

%%%%%%%%%%%%%%%%%%%%%%%%%%%%%%%%%%%%%%%
\noindent\textbf{Verifying CR/MR.}
% 活性区画内で，元画像のラベルに対応するReLU-FNN outputが，他のラベルのReLU-FNN outputよい大きければ，CRは満たされる．
When $x0$ is fixed, each activation region of the ReLU-FNN $f(g(\theta,x0)): \Theta \to Y$ is a region in the perturbation parameter space $\Theta$.
Within an activation region $\eta \subset \Theta$ of the ReLU-FNN $f(g(\theta,x0))$, $CR(f,g,x0,\eta)$ is satisfied if and only if the ReLU-FNN output corresponding to the label of the original image $x0$ cannot be less than the ReLU-FNN outputs of all other labels, i.e., the following equation holds:
$min_{j \in Y \setminus \{f(x0)\}, \theta \in \eta} f_{f(x0)}(x) - f_j(g(\theta,x0)) > 0 \Leftrightarrow CR(f,g,x0,\eta)$
Each DNN output $f_j(g(\theta,x0))$ is linear within $\eta$, and thus, the left-hand side of the above equation can be determined soundly and completely by using an LP solver (Eq.~\ref{eq:lp_encoding}-(c) in Appendix~\ref{sec:linearity}), such as \emph{scipy.optimize.linprog} \footnote{\url{https://docs.scipy.org/doc/scipy/reference/generated/scipy.optimize.linprog.html}}.
Similarly, $MR(f,g,x0,\eta)$ is satisfied if and only if the ReLU-FNN output corresponding to the label of the original image $x0$ cannot be greater than the ReLU-FNN outputs of any other labels.

%%%%%%%%%%%%%%%%%%%%%%%%%%%%%%%%%%%%%%%
\noindent\textbf{Verifying AR/IR.}
% 活性区画内で，attention inconsistencyが閾値以下なら，ARは満たされる．
Within an activation region $\eta \subset \Theta$ of the ReLU-FNN $f(g(\theta,x0))$, $AR(f,g,x0,\eta,\delta)$ is satisfied if and only if the following equation holds:
$max_{\theta \in \eta} ai(\theta,x0) \leq \delta \Leftrightarrow AR(f,g,x0,\eta,\delta)$
If $filter(\cdot)$ and $dist(\cdot)$ are both convex downward functions, as the sum of convex downward functions is also a convex downward function, the left-hand side of the above equation can be determined by comparing the values at both ends.
On the other hand, $IR(f,g,x0,\eta,\delta)$ is satisfied if and only if the following equation holds:
$min_{\theta \in \eta} ai(\theta,x0) > \delta \Leftrightarrow IR(f,g,x0,\eta,\delta)$
The left-hand side of the above equation can be determined by using an convex optimizer, such as \emph{scipy.optimize.minimize} \footnote{\url{https://docs.scipy.org/doc/scipy/reference/generated/scipy.optimize.minimize.html}}
Note that if the saliency-map is expected to keep against perturbations, the above optimization is unnecessary because $ai(\theta \in eta,x0)$ is constant.

%%%%%%%%%%%%%%%%%%%%%%%%%%%%%%%%%%%%%%%%%%%%%%%%%%%%%%%%%%%%%%%
% Experiments
%%%%%%%%%%%%%%%%%%%%%%%%%%%%%%%%%%%%%%%%%%%%%%%%%%%%%%%%%%%%%%%
\section{Experimental Evaluation} \label{sec:exp}

To confirm the usefulness of our method, we conducted evaluation experiments.

%%%%%%%%%%%%%%%%%%%%%%%%%%%%%%%%%%%%%%%
\noindent\textbf{Setups.}
Table~\ref{tab:dnns} shows the ReLU-FNNs for classification used in our experiments; each FNN uses different training data and architectures.
We inserted semantic perturbation layers (cf. Section~\ref{sec:encode_sp}) in front of each ReLU-FNN for classification during each experiment; the layers had total 1,568 neurons.
\begin{table}[t]
  \caption{The ReLU-FNNs we used in the experiments: where, ``Conv'' and ``FC'' in column Layers denote convolutional and fully connected layers, respectively. 
    % a value in column L1 denotes the strength of L1 regularization. ``B'', ``P'', and ``T'' in column DA denote semantic perturbation ``Brightness change'', ``Patch'', and ``Translation'' used for data augmentation in ReLU-FNN training, respectively.
    }
  \label{tab:dnns}
  \begin{minipage}[t]{.45\textwidth}
    \centering
    \begin{tabular}{l|rc}
      Name                       & \#Neurons & Layers                    \\
      \hline \hline
      M-FNN-100                  & 100       & FC$\times$2               \\
      M-FNN-200                  & 200       & FC$\times$4               \\
      M-FNN-400                  & 400       & FC$\times$8               \\
      M-FNN-800                  & 800       & FC$\times$16              \\
      M-CNN-S                    & 2,028     & Conv$\times$2,FC$\times$1 \\
      M-CNN-M                    & 14,824    & Conv$\times$2,FC$\times$1 \\
    \end{tabular}
  \end{minipage}
  \hfill
  \begin{minipage}[t]{.45\textwidth}
    \centering
    \begin{tabular}{c|rl}
      Name                       & \#Neurons & Layers                    \\
      \hline \hline
      F-FNN-100                  & 100       & FC$\times$2               \\
      F-FNN-200                  & 200       & FC$\times$4               \\
      F-FNN-400                  & 400       & FC$\times$8               \\
      F-FNN-800                  & 800       & FC$\times$16              \\
      F-CNN-S                    & 2,028     & Conv$\times$2,FC$\times$1 \\
      F-CNN-M                    & 14,824    & Conv$\times$2,FC$\times$1 \\
    \end{tabular}
  \end{minipage}
\end{table}

All experiments were performed on virtual computational resource ``rt\_C.small'' (with CPU 4 Threads and Memory 30 GiB) of physical compute node ``V'' (with \textit{2 CPU; Intel Xeon Gold 6148 Processor 2.4 GHz 20 Cores (40 Threads)}, and \textit{12 Memory;	32 GiB DDR4 2666 MHz RDIMM (ECC)}) in \textit{AI Bridging Cloud Infrastructure (ABCI)} \footnote{\url{https://docs.abci.ai/en/system-overview/}} provided by \textit{National Institute of Advanced Industrial Science and Technology (AIST)}.

We implemented our method in a python tool for evaluation; in addition, we will make the tool available at \url{https://zenodo.org/record/6544905} \footnote{However, due to our internal procedures, we cannot publish it until at least May 21.}.

%%%%%%%%%%%%%%%%%%%%%%%%%%%%%%%%%%%%%%%
% \subsection*{Experiments and Results.}

\begin{figure}[t]
  \centering
  \includegraphics[width=\linewidth]{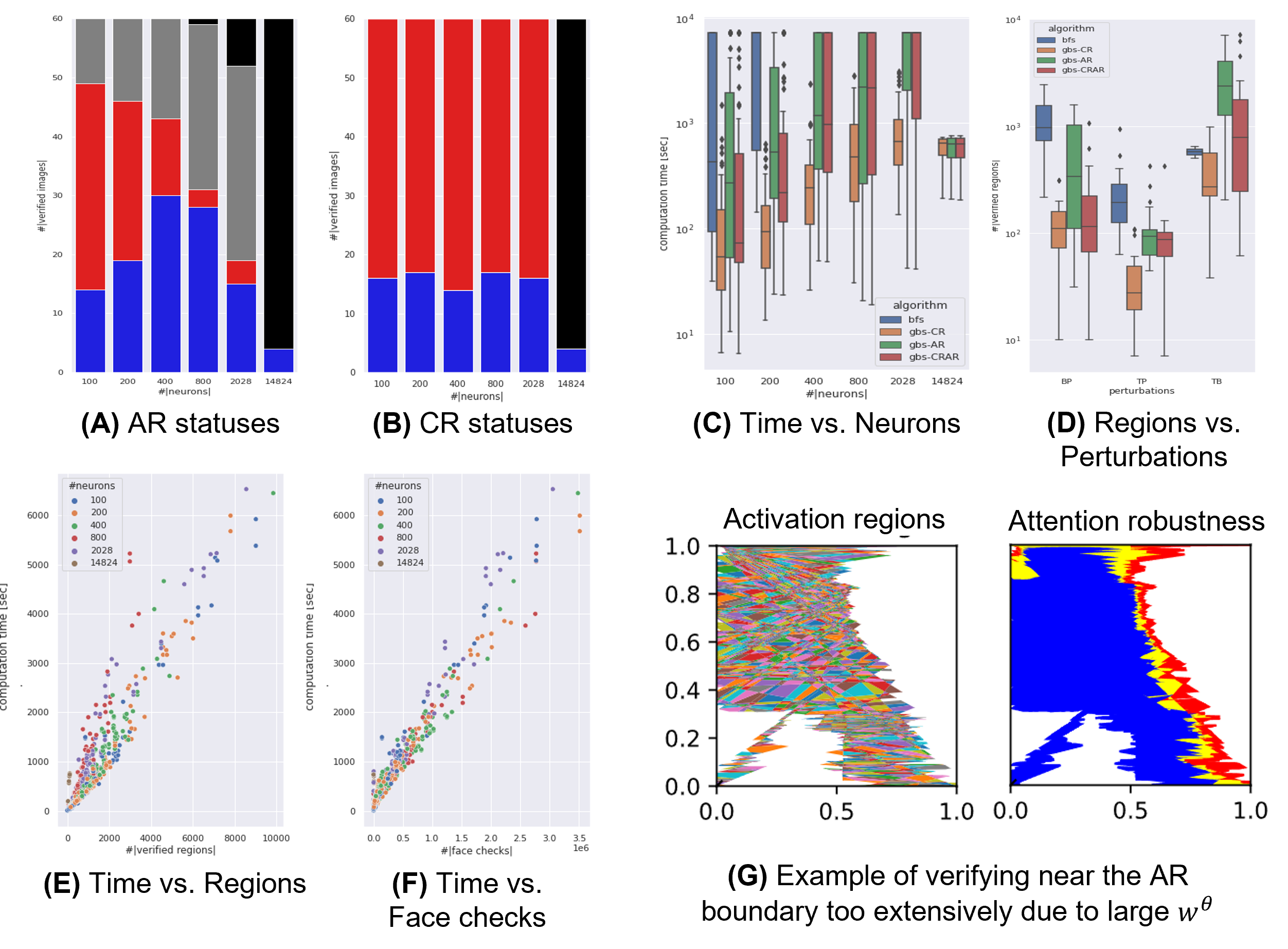}
  \caption{Experimental results.}
  \label{fig:exp_result_b}
\end{figure}

%%%%%%%%%%%%%%%%%%%%%%%%%%%%%
% Result: RQ1
%%%%%%%%%%%%%%%%%%%%%%%%%%%%%
\noindent\textbf{\RQi}\;
% 実験内容
For each ReLU-FNN, we measured the computation time of algorithms \emph{GBS} and \emph{BFS} with 10 images selected from the end of each dataset (i.e., Indexes 69990-69999. These images were not used in the training of any ReLU-FNNs).
GBS also distinguishes between \emph{gbs-CR} that traverses the CR boundary, \emph{gbs-AR} that traverses the AR boundary, and \emph{gbs-CRAR} that traverses the boundary of the regions satisfying both CR and AR.
Furthermore, we measured above per the combinations of semantic perturbations (BP), (TP), and (TB) that denote ``brightness change / patch'', ``translation / patch'', and ``translation / brightness change'', respectively.
We used the definitions $filter(x) \defeq x$, $dist(x,x') \defeq ||x-x'||_2$, $\delta \defeq 3.0$, and $w^\delta = 0.2$.

% BFSよりもGBSは早い．
Figure~\ref{fig:exp_result_b}-(C) shows the trend of increasing computation time with increasing the number of neurons for each algorithm as a box plot on a log scale.
Note that each verification timed out at 2 hours, and thus, the upper limit of computation time (y-axis) was 7,200 seconds.
We can see that algorithm BFS took more computation time than GBS.
Figure~\ref{fig:exp_result_b}-(D) shows the number of verified activation regions for each combination of perturbations; we can see that GBS reduced the number of regions to traverse compared to BFS as intended.

% ARはCRよりも時間がかかっている．
However, we can also see that gbs-AR took longer to traverse more activation regions than gbs-CR.
Figure~\ref{fig:exp_result_b}-(A) and (B) show the breakdown of each verification result for gbs-AR and gbs-CR; where, blue, red, gray, and black bars denote robust, not-robust, timed out, and out-of-memory, respectively.
The figures show that gbs-AR timed out at a higher rate for smaller size DNNs than gbs-CR.

% ニューロン数が増えると検証時間は指数的に増加する．
Moreover, we can also see that the median computation time increased exponentially with the number of neurons for all algorithms in Figure~\ref{fig:exp_result_b}-(C).
This result suggests that exact verification-based traversing is not applicable to practical-scale DNNs such as \textit{VGG16}\cite{Karen_VGG_ICLR_2015}, and fundamental scalability improvement measures, such as incorporating approximate verification, are needed.

%%%%%%%%%%%%%%%%%%%%%%%%%%%%%
% Result: RQ2
%%%%%%%%%%%%%%%%%%%%%%%%%%%%%
\noindent\textbf{\RQii}\;
% 検証した活性領域数とFaceチェック回数が計算時間の主要因．
Figure~\ref{fig:exp_result_b}-(E) and (F) show the correlation between computation time and the number of verified activation regions and the number of face checking as scatter plots, respectively: where the results of experiments that timed out are excluded.
The figures show strong positive correlations for each DNN size (the number of neurons), especially in the number of face checking; thus, reducing redundant regions and faces in understanding the boundaries should directly reduce verification time.
For example, Figure~\ref{fig:exp_result_b}-(G) shows an example of verifying activation regions near the AR boundary too extensively due to large hyperparameter $w^\theta = 0.2$.
Like gbs-CR, AR boundary also needs to be able to narrow down the search to only the regions on the AR boundaries.

%%%%%%%%%%%%%%%%%%%%%%%%%%%%%%%%%%%%%%%%%%%%%%%%%%%%%%%%%%%%%%%
% Discussion
%%%%%%%%%%%%%%%%%%%%%%%%%%%%%%%%%%%%%%%%%%%%%%%%%%%%%%%%%%%%%%%
\section{Discussion}

%%%%%%%%%%%%%%%%%%%%%%%%%%%%%%%%%%%%%%%
\noindent\textbf{Internal Validity.}
% Outermost boundaryだけでは違反領域の見逃しがありうる．
Using the outermost $CR$/$AR$ boundary as a trend of weakness to a combination of semantic perturbations, not all regions inside the outermost $CR$/$AR$ boundary satisfy the $CR$/$AR$ property.
%
% ARの閾値δには恣意性がある．
Unlike $CR$, $AR$ requires hyperparameters $\delta, w^\delta \in \mathbb{R}$; however, there are no clear criteria for setting them.

%%%%%%%%%%%%%%%%%%%%%%%%%%%%%%%%%%%%%%%
\noindent\textbf{External Validity.}
% 線形近似できない意味的摂動は検証できない．
We focus here on brightness change, patch, and translation perturbations. Still, our method applies to any semantic perturbations as long as they can be represented or approximated with sufficient accuracy.
%
% 画像の変化に対して期待する顕著性マップの変化が定義できない意味的摂動は検証できない．
Our method is not applicable to a semantic perturbation $g$ for which a corresponding saliency-map change $\tilde{g}$ cannot be defined and computed, e.g., DNNs that generate an image from an input text (such as \cite{Ramesh_DALL_ICML_2021}).
%
% CNNのMaxPool層には対応していない．
Our method has not supported the MaxPool layer of CNNs\cite{Alex_NIPS_2012}.

%%%%%%%%%%%%%%%%%%%%%%%%%%%%%%%%%%%%%%%
\subsection*{Performance Bottlenecks.}
The computation time increases exponentially with the number of ReLU-FNN neurons. Thus, our current method is not yet applicable on a practical scale, such as VGG16 \cite{Karen_VGG_ICLR_2015}.
If it is difficult to enumerate all activation regions, GBS traverses only regions on the outermost boundary. In the future, it may be better to use ReLU relaxation for dense areas.

%%%%%%%%%%%%%%%%%%%%%%%%%%%%%%%%%%%%%%%%%%%%%%%%%%%%%%%%%%%%%%%
% Related work
%%%%%%%%%%%%%%%%%%%%%%%%%%%%%%%%%%%%%%%%%%%%%%%%%%%%%%%%%%%%%%%
\section{Related Work}

%%%%%%%%%%%%%%%%%%%%%%%%%%%%%%%%%%%%%%%
\noindent\textbf{Robustness Verification.}
% 注目点の検証技術は提案されていない．
To the best of our knowledge, there have been no reports that formulate the attention robustness verification problem or that propose the method for such problem; e.g., \cite{Xiaowei_Computer_2020,Caterina_arXiv_2021,Rob_CSUR_2021}.
% DeepPolyは回転摂動を，DeepGはさらに多くの摂動を検証可能にした．
\cite{Gagandeep_DeepPoly_POPL_2019} first verified robustness against image rotation, and \cite{Mislav_DeepG_NIPS_2019} verified robustness against more semantic perturbations, such as image translation, scaling, shearing, brightness change, and contrast change. 
% しかし，我々は本稿で，分類ロバスト性が不十分な例を示した．
However, in this paper, we have demonstrated that attention robustness more accurately captures trends in weakness for the combinations of semantic perturbations than existing classification robustness in some cases.
% 近似検証は境界近傍に弱い
In addition, As \cite{Matthew_SyReNN_PLDI_2021} reported, approximated verification methods like \textit{DeepPoly}\cite{Gagandeep_DeepPoly_POPL_2019} fail to verify near the boundary. 
In contrast, our GBS enables verification near the boundary by exploratory and exact verification.

% 摂動用のDNN層を前段に挿入する工夫がある．
\cite{Jeet_Semantify_CVPR_2020} proposed that any Lp-norm-based verification tools can be used to verify the classification robustness against semantic perturbations by inserting special DNN layers that induce semantic perturbations in the front of DNN layers for classification.
% 我々も検証問題を低次元化するために採用した．
In order to transform the verification problem on the inherently high-dimensional input image space $X$ into one on the low-dimensional perturbation parameter space $\Theta$, we adopted their idea, i.e., inserting DNN layers for semantic perturbations ($\Theta \to X$) in front of DNN layers for classification ($X \to Y$).
% ただし，勾配∂f/∂xを計算する手法は我々独自である．
However, it is our original idea to calculate the value range of the gradient for DNN output ($\partial f_j(g(\theta,x_i))/\partial x_i$) within an activation region on the perturbation parameter space (cf. Sections~\ref{sec:calc_ac}-\ref{sec:verify}).

%%%%%%%%%%%%%%%%%%%%%%%%%%%%%%%%%%%%%%%
\noindent\textbf{Traversing activation regions.}
% 探索的検証手法は幾つか提案されている．
Since \cite{Matt_GeoCert_NIPS_2019} first proposed the method to traverse activation regions, several improvements and extensions have been proposed \cite{Lim_LayerCert_ICML_2020,Aymeric_FGP_ICLR_2021}.
% 既存手法はBFSで最大安全半径を算出する．
All of them use all breadth-first searches with a priority queue to compute the maximum safety radius or the maxima of the given objective function in fewer iterations.
% 本手法は飛び地を避けるためにBFSを使う．
In contrast, our algorithm GBS uses a breadth-first search with a priority queue to reach the outermost $CR$/$AR$ boundary in fewer iterations while avoiding enclaves.

% 探索時間はDNNサイズに対して指数的に増える．
\cite{Aymeric_FGP_ICLR_2021} responded to the paper reviewer that traversing time would increase exponentially with the size of a DNN \footnote{\url{https://openreview.net/forum?id=zWy1uxjDdZJ}}.
% 我々も同様の傾向を実験で示した．
Our experiment also showed that larger DNNs increase traversing time due to the denser activation regions.
% 活性区画数の増大は，探索的手法の最大の障壁になる．
The rapid increase in the number of activation regions will be one of the biggest barriers to the scalability of traversing methods, including our method.
% 活性区画数の見積もりは指数的に増大するけれど，実測は少ないことが知られている．
Although the upper bound theoretical estimation for the number of activation regions increases exponentially with the number of layers in a DNN \cite{Peter_DRLSimplex_Doctor_2021}, \cite{Boris_NIPS_2019} reported that actual DNNs have surprisingly few activation regions because of the myriad of infeasible activation patterns.
% そのため，実世界における実測値を把握する必要がある．
Therefore, it will need to understand the number of activation regions of DNNs operating in the real world.
% TODO: 実験したDNNのサイズが既存研究よりも大きいことをアピールする．

% % 並列化でスケーラビリティ向上する方向性があるけれど，本稿では出来ていない．
To improve scalability, there is the method of dividing the input space and verifying each in perfectly parallel \cite{Caterina_LIBRA_SPLASH_2020}.
However, our method has not been fully parallelized yet, because we have focused on accurately calculating the outermost $CR$/$AR$ boundary and attention robustness in this study.
% % 低次元化でスケーラビリティ向上する方向性があり，本手法でも活用している．
To improve scalability, there are several methods of targeting only low-dimensional subspaces in the high-dimensional input space for verification \cite{Matthew_ExactLine_NIPS_2019,Matthew_SyReNN_TACAS_2021,Matthew_GenProve_PLDI_2021}.
We have similarly taken advantage of low-dimensionality, e.g., using low-dimensional perturbation parameters to represent high-dimensional input image pixels as mediator variables (i.e., curried perturbation function $g^{x0}(\theta) = x'$) to reduce the elapsed time of LP solvers, determining the stability of neuron activity from few vertices of perturbation parameter space $\Theta$.

%%%%%%%%%%%%%%%%%%%%%%%%%%%%%%%%%%%%%%%
\noindent\textbf{Saliency-Map.}
% 勾配から注目画素を求める手法は多く提案されている．
Since \cite{Karen_ICLR_2014} first proposed the method to obtain a saliency-map from the gradients of DNN outputs with respect to an input image (i.e., $\partial f_j(x) / \partial x_i$), many improvements and extensions have been proposed \cite{Daniel_SmoothGrad_ICMLVIZ_2017,Mukund_IG_ICML_2017,Selvaraju_GradCAM_ICCV_2017}.
% 我々はKarenに従って，attention-mapを定式化した．
We formulated an attention-map primarily using the saliency-map definition by \cite{Karen_ICLR_2014}.
% 他の手法への対応は今後の課題．
However, it is a future work to formulate attention robustness corresponding to improvements, such as gradient-smoothing\cite{Daniel_SmoothGrad_ICMLVIZ_2017} and line-integrals \cite{Mukund_IG_ICML_2017}.

% セマンティック摂動は注目画素を変える．
It is known that semantic perturbations can significantly change the saliency-maps \cite{Gregoire_Digital_2018,Hao_CVPR_2019,Tao_ECT_AAAI_2021}.
% 我々はHaoに従って，attention inconsistencyを定式化した．
\cite{Hao_CVPR_2019} first claimed the saliency-map should consistently follow image translation and proposed the method to quantify saliency-map consistency.
We formulated attention inconsistency $ac$ primarily using the saliency-map consistency by \cite{Hao_CVPR_2019}.

% The consistency of saliency-map is used for the loss function of semi-supervised learning \cite{Hao_CVPR_2019,Tao_ECT_AAAI_2021}.
% The gradient for adversarially trained DNNs aligns well with perceptually relevant features (such as edges) of the input image \cite{Dimitris_ICLR_2019}.

%%%%%%%%%%%%%%%%%%%%%%%%%%%%%%%%%%%%%%%%%%%%%%%%%%%%%%%%%%%%%%%
% Conclusion
%%%%%%%%%%%%%%%%%%%%%%%%%%%%%%%%%%%%%%%%%%%%%%%%%%%%%%%%%%%%%%%
\section{Conclusion and Future Work}

% What we have proposed.
We, for the first time, have presented a verification method for attention robustness based on traversing activation regions on the DNN that contains layers for semantic perturbations and layers for classification.
Attention robustness is the property that the saliency-map consistency is less than a threshold value. 
We have demonstrated that attention robustness more accurately captures trends in weakness for the combinations of semantic perturbations than existing classification robustness.
% Author's belief.
Although the performance evaluation presented in this study is not yet on a practical scale, such as VGG16 \cite{Karen_VGG_ICLR_2015}, we believe that the attention robustness verification problem we have formulated opens a new door to quality assurance for DNNs.
% Future work.
We plan to increase the number of semantic perturbation types that can be verified and improve scalability by using abstract interpretation in future work.

%%%%%%%%%%%%%%%%%%%%%%%%%%%%%%%%%%%%%%%%%%%%%%%%%%%%%%%%%%%%%%%
%% Acknowledgments
%%%%%%%%%%%%%%%%%%%%%%%%%%%%%%%%%%%%%%%%%%%%%%%%%%%%%%%%%%%%%%%
% \subsubsection{Acknowledgements} Please place your acknowledgments at
% the end of the paper, preceded by an unnumbered run-in heading (i.e.
% 3rd-level heading).

% For Editage.
% We would like to thank Editage (www.editage.com) for English language editing.

%%%%%%%%%%%%%%%%%%%%%%%%%%%%%%%%%%%%%%%%%%%%%%%%%%%%%%%%%%%%%%%
%% References
%%%%%%%%%%%%%%%%%%%%%%%%%%%%%%%%%%%%%%%%%%%%%%%%%%%%%%%%%%%%%%%
%% Bibliography
\bibliographystyle{splncs04}
% \bibliography{../main} % Not working on Overleaf.com?
\bibliography{main}

%%%%%%%%%%%%%%%%%%%%%%%%%%%%%%%%%%%%%%%%%%%%%%%%%%%%%%%%%%%%%%%
%% Appendix
%%%%%%%%%%%%%%%%%%%%%%%%%%%%%%%%%%%%%%%%%%%%%%%%%%%%%%%%%%%%%%%
\def\thesection{\Alph{section}}
\section{Appendix}

%%%%%%%%%%%%%%%%%%%%%%%%%%%%%%%%%%%%%%%
\subsection{Linearity of Activation Regions} \label{sec:linearity}

\begin{figure}
  \centering
  \includegraphics[width=15em]{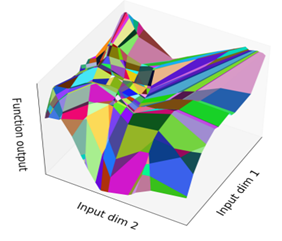}
  \caption{An example of activation regions \cite{Boris_NIPS_2019}.\\ReLU-FNN output is linear on each activation region, i.e., each output plane painted for each activation region is flat.}
  \label{fig:relu_3d_example}
\end{figure}

% 活性区画内ではReLU-FNN出力は線形．
Given activation pattern $p \in AP^f$ as constant, within activation region $ar^f(p)$ each output of ReLU-FNN $f_j(x \in ar^f(p))$ is linear for $x$ (cf. Figure~\ref{fig:relu_3d_example}) because all ReLU operators have already resolved to $0$ or $x$ \cite{Boris_NIPS_2019}.
i.e., $f_j(x \in ar^f(p)) = A'_j x + b'_j$: where, $A'_j$ and $b'_j$ denote simplified weights and bias about activation pattern $p$ and class $j$.
%
% つまり，活性区画内では勾配は定数である．
That is, the gradient of each ReLU-FNN output $f_j(x)$ within activation region $ar^f(p)$ is constant, i.e., the following equation holds: where $C \in \mathbb{R}$ is a constant value.
\begin{equation}
  \label{eq:constant_gradient}
  Feasible^f(p \in AP^f) \Rightarrow \frac{\partial f_j(x)}{\partial x_i} = C \;\; (x \in ar^f(p))
\end{equation}

% 活性区画はconvex polytopeであり，H表現できる．
An activation region can be interpreted as the H-representation of a convex polytope on input space $\mathbb{R}^{N^f}$.
Specifically, neuron activity $p_{l,n}$ and $p$ have a one-to-one correspondence with a half-space and convex polytope defined by the intersection (conjunction) of all half-spaces, because $f_n^{(l)}(x)$ is also linear when $p \in AP^f$ is constant.
% そのため，活性区画とH表現は必要に応じて相互解釈する．
Therefore, we interpret activation region $ar^f(p)$ and the following H-representation of convex polytope $HConvex^f(x;p)$ each other as needed: where, $A''$ and $b''$ denote simplified weights and bias about activation pattern $p$, and $A''_{l,n} x \leq b''_{l,n}$ is the half-space corresponding to the n-th neuron activity $p_{l,n}$ in the l-th layer.
\begin{equation}\begin{split}
  \label{eq:hconvex}
  HConvex^f(x;p) \defeq A'' x \leq b'' \;\equiv\; \bigwedge_{l,n} A''_{l,n} x \leq b''_{l,n}
\end{split}\end{equation}

%%%%%%%%%%%%%%%%%%%%%%%%%%%%%%%%%%%%%%%
\subsection{Connectivity of Activation Regions}

% Faceを反転させた活性区画は隣接している．
When feasible activation patterns $p,p' \in AP^f$ are in a relationship with each other that flips single neuron activity $p_{l,n} \in \{0,1\}$, they are connected regions because they share single face $HFace^f_{l,n}(x;p) \defeq A''_{l,n} x = b''_{l,n}$ corresponding to flipped $p_{l,n}$ \cite{Matt_GeoCert_NIPS_2019}.
% そのため，適当に反転させていくことで活性区画は探索できる．
It is possible to flexibly traverse activation regions while ensuring connectivity by selecting a neuron activity to be flipped according to a prioritization; several traversing methods have been proposed \cite{Matt_GeoCert_NIPS_2019,Lim_LayerCert_ICML_2020,Aymeric_FGP_ICLR_2021}.
% ただし，反転できない活性状態は多く存在する．
However, there are generally rather many neuron activities that become infeasible when flipped \cite{Matt_GeoCert_NIPS_2019}.
% 例の説明．
For instance, half-spaces $h_{1,3}$ is a face of activation region $\eta$ in Figure~\ref{fig:running_example_a}-(1a); thus, flipping neuron activity $p_{1,3}$, \textit{GBS} can traverse connected region $\eta$ in Figure~\ref{fig:running_example_a}-(1b).
In contrast, half-space $h_{1,1}$ is not a face of activation region $\eta$ in Figure~\ref{fig:running_example_a}-(1a); thus, flipping neuron activity $p_{1,1}$, the corresponded activation region is infeasible (i.e., the intersection of flipped half-spaces has no area).

%%%%%%%%%%%%%%%%%%%%%%%%%%%%%%%%%%%%%%%
\subsection{Hierarchy of Activation Regions}

% 活性区画は階層的であり，上流のパターンが一致する区画同士は同じ親区画に含まれる．
When feasible activation patterns $p,p' \in AP^f$ are in a relationship with each other that matches all of $L'^f$-th upstream activation pattern $p_{\leq L'^f} \defeq [ p_{l,n} \mid 1 \leq l \leq L'^f, 1 \leq n \leq N^f_l ] \;\; (1 \leq L'^f \leq L^f)$, they are included parent activation region $ar^f_{\leq L'^f}(p)$ corresponding to convex polytope $HConvex^f_{\leq L'^f}(x;p) \defeq \bigwedge_{l \leq L'^f,n} A''_{l,n} x \leq b''_{l,n}$ \cite{Lim_LayerCert_ICML_2020}.
That is, $\forall x \in ar^f(p).\; x \in ar^f_{\leq L'^f}(p)$ and $\forall x \in \mathbb{R}^{N^f}.\; HConvex^f(x;p) \Rightarrow HConvex^f_{\leq L'^f}(x;p)$.

% 同様にした，下流のパターンが一致する区画を定義できる．
Similarly, we define $L'^f$-th downstream activation pattern as $p_{\geq L'^f} \defeq [ p_{l,n} \mid L'^f \leq l \leq L^f, 1 \leq n \leq N^f_l ] \;\; (1 \leq L'^f \leq L^f)$.

%%%%%%%%%%%%%%%%%%%%%%%%%%%%%%%%%%%%%%%
\subsection{Linear Programming on an Activation Region}

Based on the linearity of activation regions and ReLU-FNN outputs, we can use \textit{Linear Programming (LP)} to compute \textbf{(a)} the feasibility of an activation region, \textbf{(b)} the flippability of a neuron activity, and \textbf{(c)} the minimum (maximum) of a ReLU-FNN output within an activation region.
We show each LP encoding of the problems (a,b,c) in the \textit{SciPy} LP form \footnote{https://docs.scipy.org/doc/scipy/reference/generated/scipy.optimize.linprog.html}: where, $p \in AP^f$ is a given activation pattern of ReLU-FNN $f$, and $p_{l,n}$ is a give neuron activity to be flipped.
\begin{equation}\begin{split}
  \label{eq:lp_encoding}
  % (a) feasibility
  \mathbf{(a)} \;\; &
    \exists x \in \mathbb{R}^{N^f}.\; HConvex^f(x;p) 
    \encode
    \min_{x}{\vec{0} x} \;\; \mathbf{s.t.,} \; A'' x \leq b'' \\
  % (b) flippability
  \mathbf{(b)} \;\;  &
    \exists x \in \mathbb{R}^{N^f}.\; HConvex^f(x;p) \land HFace^f_{l,n}(x;p) \\
    & \;\;\;\;\;\;\;\;\;\;\;\;\;\;\;\;\;\;\;\;\;\;\;\;\; \encode \min_{x}{\vec{0} x} \;\; \mathbf{s.t.,} \; A'' x \leq b'' ,\; A''_{l,n} x = b''_{l,n} \\
  % (c) minimum / maximum
  \mathbf{(c)} \;\; &
    \min_{x}{f_j(x)} \;\; \mathbf{s.t.,} \; HConvex^f(x;p)
    \encode
    \Bigl( \min_{x}{A'_j x} \;\; \mathbf{s.t.,} \; A'' x \leq b'' \Bigr) + b'_j \\
\end{split}\end{equation}

%%%%%%%%%%%%%%%%%%%%%%%%%%%%%%%%%%%%%%%
\subsection{Full Encoding Semantic Perturbations} \label{sec:encode_sp_full}

\todo{Satoshi: I will reformat this chapter, like Section~\ref{sec:encode_sp}}.
We focus here on the perturbations of brightness change (B), patch (P), and translation (T), and then describe how to encode the combination of them into ReLU-FNN $g^{x0}: \Theta \to X$: where, $|\theta^{(l)}| = \dim \theta^{(l)}$, $w$ is the width of image $x0$, $px,py,pw,ph$ are the patch x-position, y-position, width, height, and $tx$ is the amount of movement in x-axis direction. 
Here, perturbation parameter $\theta \in \Theta$ consists of the amount of brightness change for (B), the density of the patch for (P), and the amount of translation for (T).
In contrast, perturbation parameters not included in the dimensions of $\Theta$, such as $w,px,py,pw,ph,tx$, are assumed to be given as constants before verification.
\begin{equation*}\begin{split}
  g(\theta,x0) &\encode g^{x0}(\theta)       \;\; \rem{currying $g$ with given constant $x0$.} \\
  g^{x0}(\theta) &= g^{(5)}(\theta \circ x0) \;\; \rem{concat $x0$} \\
  g^{(1)}(\mu) &= A^{(T)} \mu                   \;\; \rem{translate} \\
  g^{(2)}(\mu) &= A^{(P)} g^{(1)}(\mu)          \;\; \rem{patch} \\
  g^{(3)}(\mu) &= A^{(B)} g^{(2)}(\mu)          \;\; \rem{brightness change} \\
  g^{(4)}(\mu) &= -ReLU(g^{(3)}(\mu)) + \vec{1} \;\; \rem{clip $max(0,x_i)$} \\
  g^{(5)}(\mu) &= -ReLU(g^{(4)}(\mu)) + \vec{1} \;\; \rem{clip $min(1,x_i)$} \\
\end{split}\end{equation*}
\begin{equation*}\begin{split} 
  & A^{(B)} = \left[a^{(B)}_{r,c}\right], A^{(P)} = \left[a^{(P)}_{r,c}\right], A^{(T)} = \left[a^{(T)}_{r,c}\right] \\
\end{split}\end{equation*}
\begin{equation*}\begin{split} 
  % (B) brightness change
  & a^{(B)}_{r,c} = \left\{
      \begin{array}{lll}
        1 & (c = 1 \land r \geq |\theta^{(l+1)}|) & \rem{add $\theta^{(l)}_1$} \\
        1 & (c = r + 1)                           & \rem{copy $\theta^{(l)}_{\geq 2}$ and $x_i$} \\
        0 & (otherwise) \\
      \end{array}\right. \\
% \end{split}\end{equation*}
% \begin{equation*}\begin{split} 
  % (P) patch
  & a^{(P)}_{r,c} = \left\{
      \begin{array}{lll}
        1 & (c = 1 \land On(r)) & \rem{add $\theta^{(l)}_1$} \\
        1 & (c = r + 1)         & \rem{copy $\theta^{(l)}_{\geq 2}$ and $x_i$} \\
        0 & (otherwise) \\
      \end{array}\right. \\
  & On(r) \defeq \mathbf{let\;} i \coloneqq r - |\theta^{(l+1)}|.\; (px \leq \lfloor i / w \rfloor \leq px+pw) \land (py \leq i \bmod w \leq py+ph) \\
\end{split}\end{equation*}
\begin{equation*}\begin{split} 
  % (T) translation
  & a^{(T)}_{r,c} = \left\{\begin{array}{lll}
        1                         & (c = r + 1 \land r \leq |\theta^{(l+1)}|)              & \rem{copy $\theta^{(l)}_{\geq 2}$} \\
        0                         & (c = 1 \land \lnot (1 \leq t(r) \leq s(r) \leq N))     & \rem{zero padding} \\
        x0_{tgt(r)} - x0_{src(r)} & (c = 1 \land r \geq |\theta^{(l+1)}|)                  & \rem{add $\theta^{(l)}_1 \Delta x0_i$} \\
        1                         & (c = s(r) + |\theta^{(l)}| \land r > |\theta^{(l+1)}|) & \rem{copy $x0_i$} \\
        0                         & (otherwise) \\
      \end{array}\right. \\
  & s(r) \defeq \mathbf{let\;} i \coloneqq r - |\theta^{(l+1)}|.\; (\lfloor i / w \rfloor + tx - 1) w + (i \bmod w) \\
  & t(r) \defeq \mathbf{let\;} i \coloneqq r - |\theta^{(l+1)}|.\; (\lfloor i / w \rfloor + tx - 2) w + (i \bmod w) \\
\end{split}\end{equation*}

%%%%%%%%%%%%%%%%%%%%%%%%%%%%%%%%%%%%%%%
\subsection{Images used for our experiments}

\begin{figure}[t]
  \begin{tabular}{cc}
    \begin{minipage}[t]{0.95\hsize}
      \centering
      \includegraphics[keepaspectratio, scale=0.35]{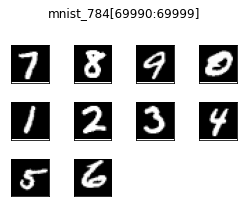}
      \caption{MNIST images used for experiments.}
      \label{fig:images_MNIST}
    \end{minipage} \\

    \begin{minipage}[t]{0.95\hsize}
      \centering
      \includegraphics[keepaspectratio, scale=0.35]{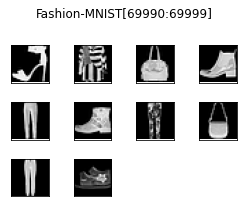}
      \caption{Fashion-MNIST images used for experiments.}
      \label{fig:images_Fashion-MNIST}
    \end{minipage}
  \end{tabular}
\end{figure}

We used 10 images (i.e., Indexes 69990-69999) selected from the end of the MNIST dataset (cf. Figure~\ref{fig:images_MNIST}) and the Fashion-MNIST dataset (cf. Figure~\ref{fig:images_Fashion-MNIST}), respectively.
We did not use these images in the training of any ReLU-FNNs.

%%%%%%%%%%%%%%%%%%%%%%%%%%%%%%%%%%%%%%%
\subsection{An example of Lemma~\ref{eq:lemma1}} \label{eg:lemma1}

Lemma~\ref{eq:lemma1} is reprinted below.
\begin{equation*}
  \frac{\partial f_j(x)}{\partial x_i} = C \;\; (x \in \{ g^{x0}(\theta) \mid \theta \in ar^{f \circ g}(p) \})
\end{equation*}

% 合成関数の偏微分からも同じことが示せる．
% This fact is confirmed by the partial derivatives of the composite function.
% Regarding $\frac{\partial f_j(g^{x0}(\theta))}{\partial \theta_s}$ is constant within activation region $ar^{f \circ g}(p)$.
% Regarding $\frac{\partial g^{x0}_i(\theta)}{\partial \theta_s}$, because $g^{x0}_i(\theta) = x_i$, $\frac{\partial x_i}{\partial \theta_s}$ is constant within parent activation region $ar^g(p_{\leq L^g})$.
% Thus, $\frac{\partial f_j(g^{x0}(\theta)) / \partial \theta_s}{\partial x_i / \partial \theta_s} = \frac{\partial f_j(x)}{\partial x_i}$ is also constant within activation region $ar^{f \circ g}(p)$, because $ar^{f \circ g}(p) \subseteq ar^g(p_{\leq L^g})$.

% \begin{proof}
%   \todo{We should prove Lemma~\ref{eq:lemma1}.}
%   \todo{As for the proof of this Lemma 1, to prove that $\forall \theta \in ar^{f \circ g}(p).; g^{x0}(\theta) \in ar^f(p_{\geq L^g+1})$ could we maybe use something like $\forall x \in ar^f(p): f^(L^{-f})(x) \in ar^f_{\geq L^{'f} +1}$?}
%   \todo{Can we say that $\forall x \in ar^f(p): f^(L^{-f})(x) \in ar^f_{\geq L^{'f} +1}$? This would then prove Lemma 1 ;)}
% \end{proof}

\begin{figure}
  \centering
  \includegraphics[width=\linewidth]{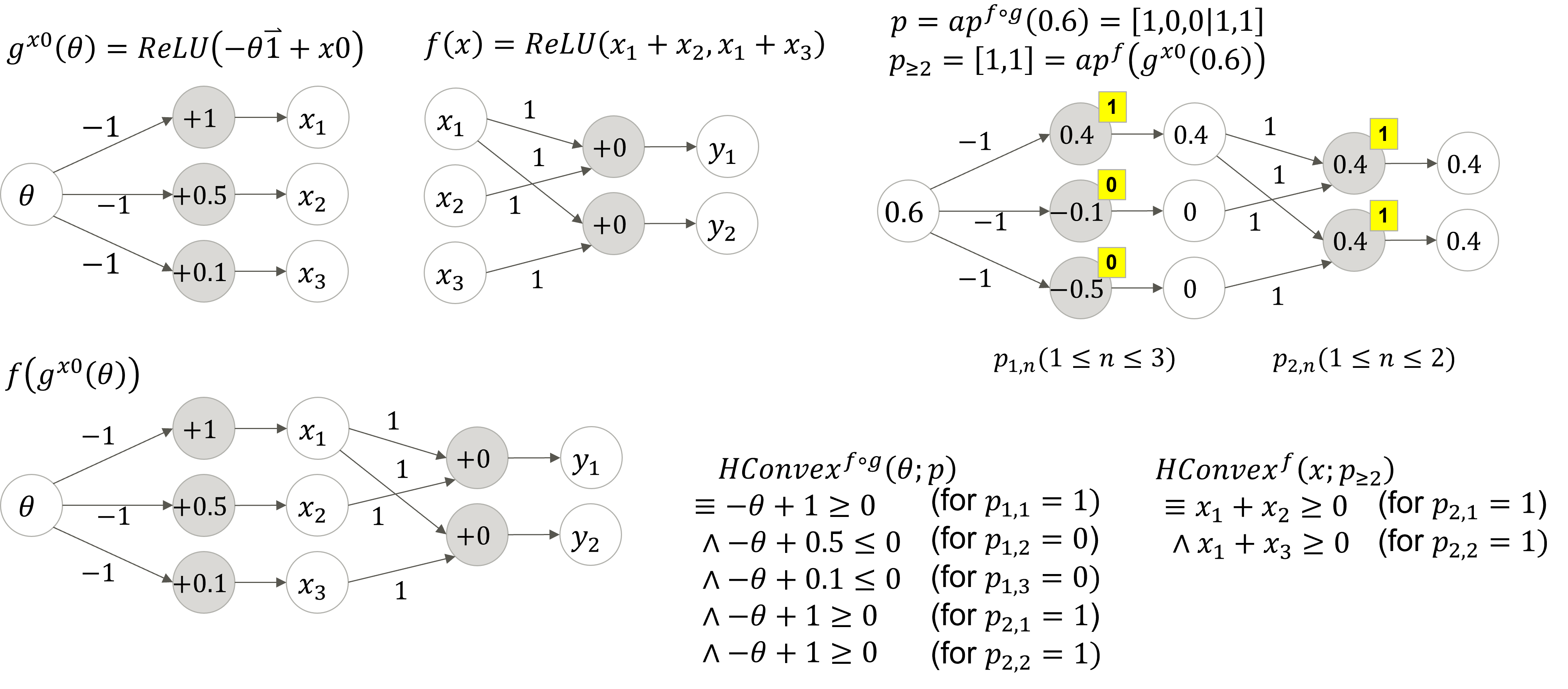}
  \caption{An image for a small example of Lemma~\ref{eq:lemma1}.}
  \label{fig:lemma1_example}
\end{figure}

\begin{itembox}{A small example of Lemma~\ref{eq:lemma1} (cf. Figure~\ref{fig:lemma1_example})}
  Let $X = [0,1]^3$, $Y = \mathbb{R}^2$, $\Theta = [0,1]^1$, $x0 \in X = (1,0.5,0.1)$, $g^{x0}(\theta \in \Theta) \in X = ReLU(-\theta\vec{1} + x0)$, and $f(x \in X) \in Y = ReLU(x_1+x_2, x_1+x_3)$.\\
  Because $g^{x0}(0.6) = ReLU(0.4,-0.1,-0.5)$ and $f(g^{x0}(0.6)) = ReLU(0.4,0.4)$, $p = ap^{f \circ g}(0.6) = [1,0,0|1,1] \in AP^{f \circ g}$.\\
  Then, $p_{\geq 2} = [1,1] = ap^f(g^{x0}(0.6)) \in AP^f$. \\
  Here, $ar^{f \circ g}(p)$ corresponding to $HConvex^{f \circ g}(\theta;p) \equiv -\theta+1 \geq 0 \land -\theta+0.5 \leq 0 \land -\theta+0.1 \leq 0 \land -\theta+1 \geq 0 \land -\theta+1 \geq 0 \equiv 0.5 \leq \theta \leq 1$, on the other hand, $ar^f(p_{\geq 2})$ corresponding to $HConvex^f(x;p_{\geq 2}) \equiv x_1+x_2 \geq 0 \land x_1+x_3 \geq 0$.\\
  Because $0 \leq x_1+x_2 = x_1+x_3 = 1-\theta \leq 0.5 \; (\theta \in ar^{f \circ g}(p))$, $\forall \theta \in ar^{f \circ g}(p).\; g^{x0}(\theta) \in ar^f(p_{\geq 2})$.
\end{itembox}

%%%%%%%%%%%%%%%%%%%%%%%%%%%%%%%%%%%%%%%
% Algorithm BFS．
%%%%%%%%%%%%%%%%%%%%%%%%%%%%%%%%%%%%%%%
\subsection{Algorithm BFS} \label{sec:bfs}

\begin{figure}[t]
  \centering
  \includegraphics[width=\linewidth]{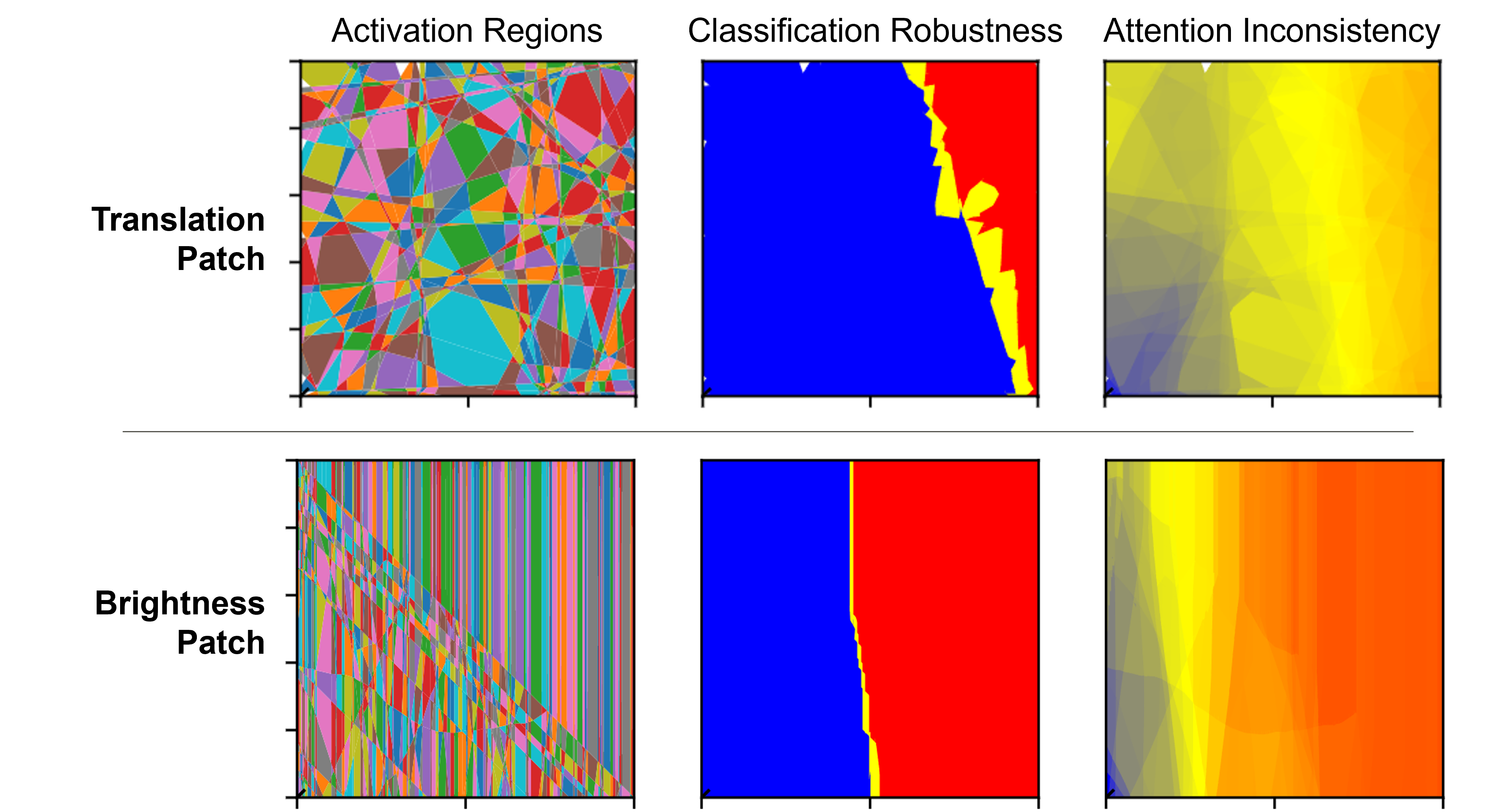}
  \caption{Examples of \textit{BFS} results. (Near the edges, polygons may fail to render, resulting in blank regions.)}
  \label{fig:bfs_regions_example}
\end{figure}

Algorithm BFS traverses entire activation regions in perturbation parameter space $\Theta$, as shown in Figure~\ref{fig:bfs_regions_example}.

Algorithm BFS initializes $Q$ with $ap^{f \circ g^{x0}}(\vec{0})$ (Line~\ref{line:bfs_init}). 
Then, for each activation pattern $p$ in $Q$ (Lines~\ref{line:bfs_while}-\ref{line:bfs_pop}), it reconstructs the corresponding activation region $\eta$ (subroutine constructActivationRegion, Line~\ref{line:bfs_car}) as the H-representation of $p$ (cf. Equation~\ref{eq:hconvex}). 
Next, for each neuron in $f \circ g^{x0}$ (Line~\ref{line:bfs_for}), it checks whether the neuron activity $p_{l,n}$ cannot flip within the perturbation parameter space $\Theta$, i.e., one of the half-spaces has no feasible points within $\Theta$ (subroutine isStable, Line~\ref{line:bfs_isStable}). 
Otherwise, a new activation pattern $p'$ is constructed by flipping $p_{l,n}$ (subroutine flipped, Line~\ref{line:bfs_flipped}) and added to the queue (Line~\ref{line:bfs_push}) if $p'$ is feasible (subroutine calcInteriorPointOnFace, Lines~\ref{line:bfs_face}-\ref{line:bfs_unface}). 
Finally, the activation region $\eta$ is simplified (Line~\ref{line:bfs_simplified}) and used to verify $CR$ (subroutine solveCR and solveVR, Lines~\ref{line:bfs_cr}-\ref{line:bfs_mr}, cf. Section~\ref{sec:verify}) and $VR$ (subroutine solveAR and solveIR, Lines~\ref{line:bfs_ar}-\ref{line:bfs_ir}, cf. Section~\ref{sec:verify}).

\begin{algorithm}[t]
  \caption{$bfs(f,g,x0,\Theta,\delta) \to (H^{CR},H^{MR},H^{CB}, H^{AR},H^{IR},H^{AB})$}
  \label{alg:bfs}

  \begin{algorithmic}[1]
    % Input/Output
    \renewcommand{\algorithmicrequire}{\textbf{Input:}}
    \renewcommand{\algorithmicensure}{\textbf{Output:}}
    \REQUIRE $f, g, x0, \Theta, \delta$
    \ENSURE  $H^{CR},H^{MR},H^{CB}, H^{AR},H^{IR},H^{AB} \subset \mathcal{P}(\Theta)$
    
    % Statements
    \STATE $H^{CR},H^{MR},H^{CB}, H^{AR},H^{IR},H^{AB} \leftarrow \{\},\{\},\{\},\{\},\{\},\{\}$
    \STATE $g^{x0} \leftarrow g(\cdot,x0)$ \rem{currying $g$ with $x0$; i.e, $g^{x0}(\theta) = g(\theta,x0)$.}
    \STATE $Q \subset AP^{f \circ g^{x0}} \leftarrow \{ap^{f \circ g^{x0}}(\vec{0})\}$ \rem{queue for breadth-first search.} \label{line:bfs_init}
    \STATE $OBS \subset AP^{f \circ g^{x0}} \leftarrow \{\}$ \rem{observed activation patterns.}

    \WHILE {$\#|Q| > 0$ \rem{loop for breadth-first search.}} \label{line:bfs_while}
      \STATE $p \leftarrow pop(Q)$  \label{line:bfs_pop}
      \STATE $OBS \leftarrow OBS \cup \{p\}$
      \STATE $\eta \leftarrow constructActivationRegion(f \circ g^{x0}, p)$ \label{line:bfs_car}

      \STATE \;\\
      \STATE \rem{Push the connected activation regions of $\eta$.}
      \STATE $FS \subset \mathbb{Z} \times \mathbb{Z} \leftarrow \{\} $ \rem{$(l,n)$ means the $n$-th neuron in $l$-th layer is a face of $\eta$}
      \FOR {$l = 1$ to the layer size of DNN $f \circ g^{x0}$, $n = 1$ to the neuron size of the $l$-th layer} \label{line:bfs_for}
        \STATE continue \textbf{if} $isStable(p,l,n,\Theta)$ \rem{skip if neuron activity $p_{l,n}$ cannot flip within $\Theta$.} \label{line:bfs_isStable}
        \STATE $p' \leftarrow flipped(p,l,n)$ \rem{flip neuron activity $p_{l,n}$.} \label{line:bfs_flipped}
        \STATE continue \textbf{if} {$p' \in OBS$} \rem{skip if $p'$ has already observed.}
        \STATE $OBS \leftarrow OBS \cup \{p'\}$

        \STATE $\theta^F \leftarrow calcInteriorPointOnFace(\eta,l,n)$ \label{line:bfs_face}
        \STATE continue \textbf{if} {$\theta^F = null$} \rem{skip if $p'$ is infeasible.} \label{line:bfs_unface}
        \STATE $FS \leftarrow FS \cup \{(l,n)\}$

        \STATE $Q \leftarrow Q \cup \{p'\}$ \rem{push.} \label{line:bfs_push}
      \ENDFOR

      \STATE \;\\
      \STATE \rem{Verify activation region $\eta$.}
      \STATE $\tilde{\eta} \leftarrow simplified(\eta,FS)$ \rem{limit the constraints on $\eta$ to $FS$.} \label{line:bfs_simplified}
      \IF {$solveCR(x0;\tilde{\eta})$} \label{line:bfs_cr}
        \STATE $H^{CR} \leftarrow H^{CR} \cup \{\tilde{\eta}\}$
      \ELSIF {$solveMR(x0;\tilde{\eta})$} \label{line:bfs_mr}
        \STATE $H^{MR} \leftarrow H^{MR} \cup \{\tilde{\eta}\}$
      \ELSE
        \STATE $H^{CB} \leftarrow H^{CB} \cup \{\tilde{\eta}\}$
      \ENDIF

      \IF {$solveAR(x0;\tilde{\eta})$} \label{line:bfs_ar}
        \STATE $H^{AR} \leftarrow H^{AR} \cup \{\tilde{\eta}\}$
      \ELSIF {$solveIR(x0;\tilde{\eta})$} \label{line:bfs_ir}
        \STATE $H^{IR} \leftarrow H^{IR} \cup \{\tilde{\eta}\}$
      \ELSE
        \STATE $H^{AB} \leftarrow H^{AB} \cup \{\tilde{\eta}^{AB}\}$
      \ENDIF
    \ENDWHILE

    \RETURN $H^{CR},H^{MR},H^{CB}, H^{AR},H^{IR},H^{AB}$
  \end{algorithmic}
\end{algorithm}

%%%%%%%%%%%%%%%%%%%%%%%%%%%%%%%%%%%%%%%
\subsection{Details of experimental results} \label{sec:exp_detail}

Table~\ref{tab:verification_statuses} shows breakdown of verification statuses in experimental results for each algorithm and each DNN size (cf. Section~\ref{sec:exp}).
In particular, for traversing AR boundaries, we can see the problem that the ratio of ``Timeout'' and ``Failed (out-of-memory)'' increases as the size of the DNN increases.
This problem is because gbs-AR traverses more activation regions by the width of the hyperparameter $w^\delta$ than gbs-CR. 
It would be desirable in the future, for example, to traverse only the small number of activation regions near the AR boundary.

\begin{table}
  \caption{Breakdown of verification statuses. ``Robust'' and ``NotRobust'' mean algorithm found ``only robust regions'' and ``at least one not-robust region'', respectively. ``Timeout'' and ``Failed'' mean algorithm did not finish ``within 2 hours'' and ``due to out-of-memory'', respectively.}
  \label{tab:verification_statuses}
  \centering
  \begin{tabular}{lr|rrrr}
    algorithm & \#neurons & Robust & NotRobust & Timeout & Failed \\
    \hline \hline
         bfs &      100 &     13 &        22 &      25 &      0 \\
         bfs &      200 &     11 &        15 &      34 &      0 \\
    \hline
      gbs-CR &      100 &     16 &        44 &       0 &      0 \\
      gbs-CR &      200 &     17 &        43 &       0 &      0 \\
      gbs-CR &      400 &     14 &        46 &       0 &      0 \\
      gbs-CR &      800 &     17 &        43 &       0 &      0 \\
      gbs-CR &     2028 &     16 &        44 &       0 &      0 \\
      gbs-CR &    14824 &      4 &         0 &       0 &     56 \\
    \hline
      gbs-AR &      100 &     14 &        35 &      11 &      0 \\
      gbs-AR &      200 &     19 &        27 &      14 &      0 \\
      gbs-AR &      400 &     30 &        13 &      17 &      0 \\
      gbs-AR &      800 &     28 &         3 &      28 &      1 \\
      gbs-AR &     2028 &     15 &         4 &      33 &      8 \\
      gbs-AR &    14824 &      4 &         0 &       0 &     56 \\
    \hline
    gbs-CRAR &      100 &     14 &        41 &       5 &      0 \\
    gbs-CRAR &      200 &     19 &        33 &       8 &      0 \\
    gbs-CRAR &      400 &     30 &        14 &      16 &      0 \\
    gbs-CRAR &      800 &     28 &         6 &      26 &      0 \\
    gbs-CRAR &     2028 &     15 &         8 &      32 &      5 \\
    gbs-CRAR &    14824 &      4 &         0 &       0 &     56 \\
  \end{tabular}
\end{table}

%%%%%%%%%%%%%%%%%%%%%%%%%%%%%%%%%%%%%%%%%%%%%%%%%%%%%%%%%%%%%%%%%%%%%%%%%%%%%%%
\end{document}